\documentclass[10pt,journal,compsoc]{IEEEtran}
\usepackage{xr}
\usepackage{soul}
\usepackage{color}
\usepackage{xurl}
\renewcommand{\hl}[1]{#1} 

\usepackage{amsmath}
\usepackage{amssymb}
\usepackage{caption}
\usepackage{subcaption}
\usepackage{graphicx}
\usepackage[export]{adjustbox}
\usepackage{dsfont}
\usepackage{array, booktabs, makecell}
\usepackage[normalem]{ulem}
\useunder{\uline}{\ul}{}

\ifCLASSOPTIONcompsoc
  \usepackage[nocompress]{cite}
\else
  \usepackage{cite}
\fi
\ifCLASSINFOpdf
\else
\fi

\def\plotexplain{\hl{Bar height indicates the average relative accuracy (relative to the best classical model for each dataset)} of all experiments with a given configuration. Gray line indicates the standard error. The \hl{horizontal} stacked bar chart labelled \textit{Wins} indicates for how many datasets each configuration gives the highest relative accuracy.}
\begin{document}
%
\title{\hl{Dual input stream transformer for vertical drift correction in eye-tracking reading data}}

%
%
%
%

\author{Thomas~M.~Mercier,~Marcin~Budka,~Martin~R.~Vasilev,~Julie~A.~Kirkby,~Bernhard~Angele,~Timothy~J.~Slattery
\IEEEcompsocitemizethanks{

\IEEEcompsocthanksitem Thomas M. Mercier, Julie~A.~Kirkby, and ~Timothy~J.~Slattery are with the Department
of Psychology, Bournemouth University, Poole,
Dorset, BH12 5BB, UK.
E-mail: tmercier@bournemouth.ac.uk
\IEEEcompsocthanksitem Bernhard~Angele is with the Nebrija Research Centre in Cognition (Centro de Investigación Nebrija en Cognición, CINC), Universidad Antonio de Nebrija, Calle de Asura, 90, 28043 Madrid, Spain and the Department of Psychology, Bournemouth University.
\IEEEcompsocthanksitem 
Marcin Budka is with the Department of Computing \& Informatics, Bournemouth University, Poole,
Dorset, BH12 5BB, UK.
\IEEEcompsocthanksitem Martin~R.~Vasilev is with the Department of Experimental Psychology, University College London, Gower Street, London, WC1E 6BT UK.
}
\thanks{This work has been submitted to the IEEE for possible publication. Copyright may be transferred without notice, after which this version may no longer be accessible.
	
© 2024 IEEE. Personal use of this material is permitted. Permission
from IEEE must be obtained for all other uses, in any current or future
media, including reprinting/republishing this material for advertising or
promotional purposes, creating new collective works, for resale or
redistribution to servers or lists, or reuse of any copyrighted
component of this work in other works.}}

%
%

\markboth{Transactions on pattern analysis and machine intelligence}%
{Shell \MakeLowercase{\textit{et al.}}: Bare Demo of IEEEtran.cls for Computer Society Journals}
%



\IEEEtitleabstractindextext{%
\begin{abstract}
We introduce a novel Dual Input Stream Transformer (DIST) for the challenging problem of assigning fixation points from eye-tracking data collected during passage reading to the line of text that the reader was actually focused on. This post-processing step is crucial for analysis of the reading data due to the presence of noise in the form of vertical drift. We evaluate DIST against \hl{eleven} classical approaches on a comprehensive suite of nine diverse datasets. \hl{We demonstrate that combining multiple instances of the DIST model in an ensemble achieves high accuracy across all datasets. Further combining the DIST ensemble with the best classical approach yields an average accuracy of 98.17~\%.} Our approach presents a significant step towards addressing the bottleneck of manual line assignment in reading research. Through extensive analysis and ablation studies, we identify key factors that contribute to DIST's success, including the incorporation of line overlap features and the use of a second input stream. Via rigorous evaluation, we demonstrate that DIST is robust to various experimental setups, making it a safe first choice for practitioners in the field.

\end{abstract}

\begin{IEEEkeywords}
	Machine Learning, Psychology, Pattern Recognition, Artificial Intelligence, Computer vision.
\end{IEEEkeywords}}

\maketitle

\IEEEdisplaynontitleabstractindextext

%
\IEEEpeerreviewmaketitle

\IEEEraisesectionheading{\section{Introduction}\label{sec:Introduction}}
\IEEEPARstart{T}{he} ability to read is an indispensable skill in modern society, making reading a prominent subject in the psychological and cognitive sciences. Eye-tracking technology has emerged as a valuable tool for uncovering the cognitive processes involved in reading, offering unique insights into the reading patterns of individuals. It involves measuring the gaze position on a computer screen over time with a typical sampling frequency of 1000 Hz \hl{(although higher and lower sampling rates such as 500 Hz and 2000 Hz are not unusual)} by analyzing the relative position of the pupil and corneal reflection centers~\cite{huttonEyeTrackingMethodology2019,raatikainenDetectionDevelopmentalDyslexia2021}. This can provide critical information about where and for how long an individual's eyes are fixating while they navigate text, revealing essential clues about the mental processes at play~\cite{rayner35thSirFrederick2009}. From the gaze position measurements, each gaze point can be considered to either be part of a fixation, \hl{a state of oculomotor control where the gaze position is held fairly constant while information is extracted from the image on the retina, or part of a saccade, which is a very fast, ballistic eye movement that moves the center of the gaze to a different part of the visual field. Depending on the properties of the visual stimulus and the task, the mean fixation duration can vary between 180 and 330~ms (with typical reading fixations having lengths between 200 and 250~ms). In some tasks, fixations as short as 50~ms and as long as 600~ms have been observed. The duration of a saccade is directly related to the distance it covers in the visual field. In reading, saccade duration usually does not exceed 30-50~ms\mbox{\cite{rayner35thSirFrederick2009}}.}

Eye movement data have proven invaluable in the study of reading behavior and in the diagnosis and treatment of specific disorders such as dyslexia and schizophrenia~\cite{jankovicBiomarkerbasedApproachesDyslexia2022,diasNeurophysiologicalOculomotorComputational2021,raatikainenDetectionDevelopmentalDyslexia2021}. By examining eye movements during reading tasks, researchers can better understand the cognitive mechanisms engaged in the comprehension of written material and potentially improve interventions for those who struggle with reading due to neurological differences or other factors. Eye-tracking experiments can also offer insight into accessibility related issues of visual stimuli such as websites~\cite{eraslanWebUsersAutism2019}. 

While the technological advances in recent decades have enabled the recording of gaze position during reading with high accuracy, eye-tracking data from such recordings consists of the raw gaze position samples and thus still requires post-processing to identify which gaze positions are part of the fixations and which are part of a saccade. Furthermore, these fixations need to be assigned to an area of interest in the reading stimulus, which can be a particular character or word, depending on the design of the experiment. 

\hl{Modern eye-trackers can record gaze position with high precision, but they can only ever be as good as their calibration. Even small, involuntary participant movements can affect the precision of the calibration, which can result in dynamically changing offset in both the horizontal and the vertical coordinate. Due to the way printed text is organized in lines, an offset in the horizontal coordinate will lead to a fixation being assigned to a different letter, or at worst an adjacent word on the same line. Because of this, horizontal drift is usually not corrected. However, a vertical offset can lead to a fixation appearing to be on an incorrect line, possibly many words away from the correct fixation location \mbox{\cite{carrAlgorithmsAutomatedCorrection2022}}. This offset can be so pronounced as to make the recorded fixation appear to lie on one or several lines above or below the line that the reader was actually focusing on. An example for this phenomenon is shown in the supplementary information.} \hl{The adjustment of the vertical coordinate of a fixation point to correct for this vertical drift is referred to as line assignment since the vertical coordinate is set to the center of the line that the fixation is assigned to. }

 \hl{Single line experiments avoid the issues associated with line assignment as assigning fixations is trivial when there is only one line, however, real-world readers do not typically read single lines on an otherwise blank page or screen. In order to get a more naturalistic sample of real-world reading}, many experiments involve full passages of text that require each fixation to be assigned to a line of text to carry out further analysis. Since simply assigning a fixation point to whichever line it is closest to can lead to incorrect assignments, more careful approaches to line assignment have to be utilized. This can present a significant hurdle to carrying out large number of trials for studies involving multi-line passages of text as the process is often carried out manually. \hl{See for example \mbox{\cite{adedejiChildrenReadingSublexical2023,scherrTextMattersEye2016,vasilevReadersUseCharacter2021a}}}.

Line assignment \hl{of fixations} is made significantly more difficult \hl{due to} noise in the tracking data, as can arise from loss of calibration of the eye-tracker, subtle head or body movements or even pupil dilation during the experiment~\cite{carrAlgorithmsAutomatedCorrection2022}. Such noise can take the form of dynamically changing vertical drift of the recorded fixations, which causes fixations to be recorded as falling on lines above or below of the actually fixated line. There have been several attempts to create algorithms to automate the line assignment process and therefore enable researchers to carry out larger studies of multi-line reading. See Carr et al.~\cite{carrAlgorithmsAutomatedCorrection2022} for a review of several available algorithms. 

\hl{Such techniques, however, have not been evaluated on multiple datasets and are commonly not easily accessible to researchers in the field of psychology. This may lead researchers to have concerns about their accuracy and reliability as evidenced by the fact that line assignment to correct vertical drift is still commonly done manually\mbox{\cite{adedejiChildrenReadingSublexical2023,scherrTextMattersEye2016,vasilevReadersUseCharacter2021a}} despite the availability of line assignment algorithms}. Hence, manual correction is considered the gold standard for addressing noise in eye-tracking fixation data~\cite{carrAlgorithmsAutomatedCorrection2022}. \hl{Unfortunately, any form of manual assignment depends on the individual carrying out the task and can yield inconsistent results based on the individual carrying out the task\mbox{\cite{hoogeHumanClassificationExperienced2018}}. Nevertheless, if the available algorithms prove insufficient, they are left with no choice but to resort to a manual approach. In reading research involving eye-tracking, this is still common~\mbox{\cite{adedejiChildrenReadingSublexical2023,scherrTextMattersEye2016,vasilevReadersUseCharacter2021a}}}. Since this is both time-consuming and increases subjectivity in the resulting assignments, it is desirable for this process to be reliably automated, an idea initially put forward by Cohen~\cite{cohenSoftwareAutomaticCorrection2013a}.

In this paper we present a novel way of tackling the line assignment problem by utilizing a deep learning (DL) architecture trained using a rank-consistent ordinal regression loss function. Our proposed architecture uses two input streams to incorporate information about \hl{both} the eye-tracking fixations \hl{and} the stimulus text used in the trial. We call our model Dual Input Stream Transformer (DIST)\footnote{\hl{Our code and the user interface can be found here:} \url{https://github.com/Gittingthehubbing/DIST-Dual_Input_Stream_Transformer}}. We show that \hl{combining an ensemble of our proposed model with classical algorithms in a "Wisdom of the Crowds" (WOC) approach} outperforms the best classical algorithms \hl{(including a WOC of all classicals) on all} datasets. \hl{The idea of combining the algorithms this way is inspired by \mbox{\cite{carrAlgorithmsAssigningFixations}}. We acknowledge that our model is unlikely to perform well for data with very different fixation patterns than those of the datasets used to train the model, such as experiments using different fonts or font sizes, for example. The datasets considered in this study all used either the Consolas or the Courier New fonts with font sizes ranging from 11 to 22. Nevertheless, the diversity of these datasets and the data normalization schemes employed should allow the model to generalize to many unseen datasets without the user having to carry out any manual line assignment or fine-tuning of the model. Furthermore, the combination with the classical models via the WOC approach should further increase resilience to data that is very unlike the training data.} \hl{We further acknowledge that due to the increased complexity of our method how the results came to be can not be easily traced. However, in practice, this does not present a significant drawback since, regardless of the algorithm used, some visual inspection of the results will be necessary. Cases where the correction algorithm made particularly large adjustments or where there is strong disagreement between different correction methods should be inspected in particular. In practice, this will only apply to a small subset of trials.} 

To the best of our knowledge \hl{this paper presents the }first comprehensive comparison of \hl{multiple} line assignment algorithms across diverse datasets in the domain of eye-tracking research. For practitioners in the field of psychology the best model will likely be the one that can be robustly applied to data from various studies. Ultimately, widespread adoption of algorithmic tools could lead to increased efficiency and consistency in the analysis of such data, fostering more reliable and accurate scientific insights.

Our contributions are as follows: 1)~introduction of a novel transformer-based architecture for the problem of line assignment of fixation coordinates, 2)~\hl{comparative} evaluation of the newly introduced \hl{approaches} on diverse \hl{data} from multiple studies, 3)~increased robustness and accuracy of line assignment to enable reading researchers to carry out larger studies involving multi-line reading experiments.

\section{Related work}
\label{sec:relwork}

Carr et al.~\cite{carrAlgorithmsAutomatedCorrection2022} recently reviewed and evaluated several \hl{classical} algorithms for correcting vertical drift in eye-tracking data. They summarized the \hl{approaches} into ten categories and implemented each approach to enable direct comparison. They reported high accuracies for most algorithms for their small dataset of manually corrected experimental data as well as synthetic data from simulated experiments. By implementing all algorithms with a similar interface, the authors greatly improved usability and comparability of the published algorithms. We adapt their implementations to compare our model to classical work. \hl{Note that while it is not discussed in their publication, Carr et al. nevertheless offers an implementation of the \textit{slice} (described below) algorithm. This brings the total number of compared classical algorithms to eleven.} As will be shown in \hl{Table~\mbox{\ref{tbl-classicCompare}} of }Section~\ref{sec:results}, the \hl{compared} algorithms showed large variability \hl{for the different dataset}. When the accuracies across all trials and datasets are averaged, \hl{the \textit{cluster}\mbox{\cite{sascha2schroederPopEye2022,carrAlgorithmsAutomatedCorrection2022}}, \textit{merge} and \textit{regress}~\mbox{\cite{carrAlgorithmsAutomatedCorrection2022,sakoeDynamicProgrammingAlgorithm1978,vintsyukSpeechDiscriminationDynamic1968}} algorithms show the best performance among the individual classical methods.}

\hl{Carr's \textit{cluster}~\mbox{\cite{sascha2schroederPopEye2022,carrAlgorithmsAutomatedCorrection2022}} implementation applies k-means clustering in order to assign all fixations to one of m clusters, where m is equal to the number of lines in the passage. The mean value of the y-coordinate of all fixations in one cluster is used to find the line number that a particular cluster corresponds to.}

\hl{The \textit{merge} algorithm as implemented by Carr~\mbox{\cite{carrAlgorithmsAutomatedCorrection2022}} uses Spakov et al.'s~\mbox{\cite{spakovImprovingPerformanceEye2019}} post hoc correction approach with modifications. It generates progressive sequences of consecutive fixations and merges them into larger sequences until there's one sequence per line of text. A y threshold defines proximity, and a regression-based merge process ensures similar gradients and low errors between merged sequences. The algorithm follows Spakov et al.'s~\mbox{\cite{spakovImprovingPerformanceEye2019}} four phases for relaxing criteria, while the number of sequences is reduced to match text lines in positional order.
}

The implementation by Carr et al.~\cite{carrAlgorithmsAutomatedCorrection2022} of the \textit{regress} algorithm is based on the R package \textit{FixAlign}~\cite{cohenSoftwareAutomaticCorrection2013a}. It fits a number of regression lines\hl{, meaning lines going through a set of points that are found by minimizing the y-distance of all points to the line,} to the unordered fixation points and subsequently assigns each fixation to its highest likelihood regression line. This identifies outliers and assigns each fixation to their most appropriate line of text.

The recently published \textit{slice} algorithm works in three main steps~\cite{glandorfSliceAlgorithmAssign2021}. Firstly, it finds fixation sequences that are likely to belong to the same line. Secondly, it goes through the sequences, starting from the longest and finds which of the sequences belong to the same or an adjacent line. Lastly, it ensures that the number of detected sequences is equal to the number of lines in the text. They report 95.3~\% accuracy for a subset of the MECO dataset.

To the best of our knowledge this paper is the first DL-based solution to the line assignment problem. Eye-tracking data have been utilized to train machine learning (ML) models to diagnose reading difficulties. \hl{While a review of the many ML related eye-tracking applications is beyond the scope of this paper, we would like to briefly discuss some examples.} In an early study by Rello and Ballesteros~\cite{relloDetectingReadersDyslexia2015} the authors used the eye-tracking data of 97 participants, 48 of which had a dyslexia diagnosis, to extract features that were deemed relevant to the problem of reading difficulty diagnosis. These features were fed into a polynomial Support Vector Machine (SVM) to classify each participant as either dyslexic or not. Using this approach the authors report a 10-fold cross-validation accuracy of 80.18~\%.

In a more recent study, Vajs et al.~\cite{vajsDyslexiaDetectionChildren2022} used a DL approach based on a convolutional neural network~(CNN) to perform dyslexia classification for a dataset of 30 subjects, half of which had a diagnosis of dyslexia. They use the gaze coordinates without extracting fixations or other features, however, they do clean the data by removing eye blinks and data points they consider invalid due to the gaze not being registered by the measurement device. After splitting each trial data into a number of sequences, the authors visualize the gaze trace with the distance between subsequent points being encoded as color. This visualization serves as the input to the CNN with each visualization being assigned the label of belonging to a dyslexic subject or not. They report an accuracy of 87~\%. These reports highlight the importance of eye-tracking data for diagnosis of widespread pathologies.

\hl{Additionally, modeling of eye-tracking data has also been used to assess second language proficiency by scoring how similar the gaze patterns, as represented by eye-movement features, of the subjects to those of native speakers and predicting scores of other language tests via a regression model~\mbox{\cite{berzakAssessingLanguageProficiency2018a}}. Furthermore, DL-based models using only features extracted from eye-movements or models that combine them with linguistic features of the text have been applied to the task of estimating reading comprehension~\mbox{\cite{ahnPredictingReadingComprehension2020,reichInferringNativeNonNative2022}}. ML models have also been used to directly predict the fixations using an approach that incorporates both the sequence of fixations and the sequence of words making up the stimulus text~\mbox{\cite{dengEyettentionAttentionbasedDualSequence2023a}}.}

\section{Datasets}
\label{sec:dsets}

We make use of data from nine studies (see Table~\ref{tbl-datasets})\footnote{\hl{The preprocessed datasets and links to the raw data, where available, can be found here can be found here: }\url{https://osf.io/zt9gn}.}. For all these studies the authors carried out manual line assignment of all fixations. \hl{In total these datasets contain 15,446 trials, although the two largest datasets, ChA and TexF, only have two lines of text in their stimulus material. The low number of lines makes the correction task for these datasets much easier than for typical paragraph data and therefore pushes up the average accuracies for all compared approaches.} \hl{Here,} one trial is considered to be the result of a participant reading one screen of text and \hl{yielding one} fixation sequence. For each of the fixations in such a sequence the ground truth consists of the index of the line to which the fixation is assigned, with this assignment being based on the gold standard human-corrected manual approach. \hl{Note that since none of the classical algorithms can discard fixations we only consider those fixations that had been assigned to a line by the human labeler, hence fixations that the human labeler discarded are not considered. Across datasets this was the case for an average of 2.8~\% (ranging from 0.15 to 8.2~\%) of fixations.} Since the stimulus texts varied in length and line counts, this results in a differing number of target classes for each dataset. Please see Section~\ref{sec:framework} for how this is handled in the model design.

\begin{table*}[t]

\centering
\caption{\hl{Characteristics and associated references for all datasets. Mean no. of fixations is how many fixations appear in a trial on average. Mean trial time is how much time in seconds passes between start and end of a trial, on average. Min and Max no. of lines gives what range of number of lines in a trial exists for a dataset, thereby giving an indication of paragraph length for each dataset. Max/Min no. of words gives the largest/smallest number of words appearing in the stimuli of the dataset. No. of trials is the total number of trials in the dataset. Mean fixation duration gives the duration of a fixation in ms. Mean no. of words per line gives the number of words per line for the trials in a dataset, on average. Mean line width gives the width of an average line in pixels. Min/Max line width gives the smallest/largest line width in pixels for the trials in a dataset. Source lists the associated publication.}}
\label{tbl-datasets}
    \begin{tabular}{cccccccccc} \hline
    \textbf{Dataset}                     & \textbf{Abbr}                                              & \textbf{ChA}                                                 & \textbf{OffN}                                                 & \textbf{TexF}                                          & \textbf{CD}                                                     & \textbf{OZ}                                                           & \textbf{Harry}                                                & \textbf{MECOde}                                                                                        & \textbf{Carr}                                                 \\ \hline
    \textbf{Mean no. of fixations}          & 143.11                                                     & 26.01                                                        & 114.16                                                        & 19.04                                                  & 99.45                                                           & 90.19                                                                 & 90.39                                                         & 247.09                                                                                                 & 208.13                                                        \\
    \textbf{Mean trial time \hl{(s)}}             & 39.49                                                      & 10.40                                                        & 31.90                                                         & 4.74                                                   & 26.84                                                           & 22.42                                                                 & 27.79                                                         & 59.42                                                                                                  & 49.43                                                         \\
    \textbf{Min no. of lines}                   & 9                                                          & 2                                                            & 1                                                             & 2                                                      & 6                                                               & 8                                                                     & 10                                                            & 10                                                                                                     & 10                                                            \\
    \textbf{Max no. of lines}                   & 12                                                         & 2                                                            & 12                                                            & 2                                                      & 8                                                               & 10                                                                    & 10                                                            & 14                                                                                                     & 13                                                            \\
    \textbf{Min no. of words}                   & 95                                                         & 15                                                           & 10                                                            & 13                                                     & 84                                                              & 82                                                                    & 133                                                           & 144                                                                                                    & 104                                                           \\
    \textbf{Max no. of words}                   & 132                                                        & 25                                                           & 155                                                           & 34                                                     & 116                                                             & 142                                                                   & 153                                                           & 225                                                                                                    & 168                                                           \\
    \textbf{No. of trials}                  & 1599                                                       & 2682                                                         & 1506                                                          & 6397                                                   & 1150                                                            & 1116                                                                  & 300                                                           & 648                                                                                                    & 48                                                            \\
    \textbf{\hl{Mean fixation duration (ms)}} & 237.15                                                     & 376.24                                                       & 208.52                                                        & 206.72                                                 & 220.63                                                          & 209.97                                                                & 234.26                                                        & 203.83                                                                                                 & 223.07                                                        \\
    \textbf{\hl{Mean no. of words per line}}     & 10.97                                                      & 8.78                                                         & 11.39                                                         & 10.10                                                  & 13.02                                                           & 10.88                                                                 & 12.54                                                         & 14.36                                                                                                  & 11.45                                                         \\
    \hl{\textbf{Mean line width (pixels)}}    & 826.50                                                     & 671.85                                                       & 964.51                                                        & 759.32                                                 & 821.42                                                          & 704.83                                                                & 1035.12                                                       & 1557.36                                                                                                & 1058.32                                                       \\
    \hl{\textbf{Max line width (pixels)}}     & 975                                                        & 800                                                          & 1316                                                          & 1376                                                   & 913                                                             & 1080                                                                  & 1106                                                          & 1770                                                                                                   & 1176                                                          \\
    \hl{\textbf{Min line width (pixels)}}     & 65                                                         & 288                                                          & 56                                                            & 372                                                    & 44                                                              & 48                                                                    & 238                                                           & 270                                                                                                    & 104                                                           \\
    \textbf{Source}                      & \cite{adedejiReturnsweepSaccadesOral2022} & \cite{adedejiChildrenReadingSublexical2023} & \cite{goldenbergMeasuredPerceivedImpact2022} & \cite{vasilevReadersUseCharacter2021} & \cite{vasilevReadingDisruptedIntelligible2019} & \cite{slatteryEyemovementExplorationReturnsweep2019} & \cite{seymourUsingEyeTrackingTechnology2022} & \cite{siegelmanExpandingHorizonsCrosslinguistic2022,glandorfSliceAlgorithmAssign2021} & \cite{carrAlgorithmsAutomatedCorrection2022} \\ 
    \hl{\textbf{Is raw data public?}}          & Yes                                                        & No                                                           &               No                                                & Yes                                                    & Yes                                                             & Yes                                                                   &                                                              No &                                                                             No                           & Yes \\ \hline 
    \end{tabular}
\end{table*}

Table~\ref{tbl-datasets} \hl{shows the dataset characteristics} which varied in a variety of aspects. This provided the model a broad basis for learning how to assign fixations to lines under varying experimental conditions. On average, the text stimuli varied between 1 and 14 lines yielding on average 19 to 247 fixations per trial. The average fixation duration is fairly consistent except for the ChA dataset having nearly twice the average duration. All datasets except MECOde and Carr were collected at Bournemouth University in the UK. The MECOde dataset was kindly shared by the creators of the MECO dataset~\cite{siegelmanExpandingHorizonsCrosslinguistic2022,glandorfSliceAlgorithmAssign2021}. The Carr dataset is publicly available and associated with Carr et al.~\cite{carrAlgorithmsAutomatedCorrection2022}.

In addition to the datasets available from eye-tracking studies, we produce a synthetic dataset of fixation sequences with different kinds of noise added to them. To produce the synthetic sequences we adapt code from~\cite{carrAlgorithmsAutomatedCorrection2022}. \hl{Based on the \emph{wikitext} dataset~\mbox{\cite{merityPointerSentinelMixture2016}} we generate passages consisting of 8 to 14 lines of English text with each line being up to 130 characters long and the line height varying between 49 and 79 pixels. The choice for each parameter for each passage was based on a uniform random distribution. For each passage a sequence of up to 500 fixations is generated. To simulate the effects of loss of equipment calibration and other disturbances that can result in both random and systematic errors in the recorded fixation positions the y-coordinate is determined by $y=\mathcal{N}(l_{y},d_{{{\mathrm{noise}}}})+l_{y}d_{\mathrm{shift}}$, where $\mathcal{N}$ is the normal distribution, $l_y$ is the y-center of the line, $d_{{{\mathrm{noise}}}}$ is the standard deviation of noise and $d_{\mathrm{shift}}$ is the y-shift in pixels.} To increase how realistic the synthetic fixation patterns are, within-line and between-line regressions are added. Regressions refer to the phenomenon of readers fixating parts of the text that lie before the current position. \hl{Please see the supplementary material and }\cite{carrAlgorithmsAutomatedCorrection2022} for details on how this is implemented.

\hl{While there are DL-based methods to produce fixation sequences~\mbox{\cite{dengEyettentionAttentionbasedDualSequence2023a}}, we focus on Carr's approach due to its ability to produce a large set of sequences with the ability to control both the type and severity of noise as well as the corresponding line correction for each fixation.}

\begin{figure}
	\centering
	\begin{subfigure}{0.99\linewidth}
		\centering 
		\includegraphics[width=1.0\linewidth]{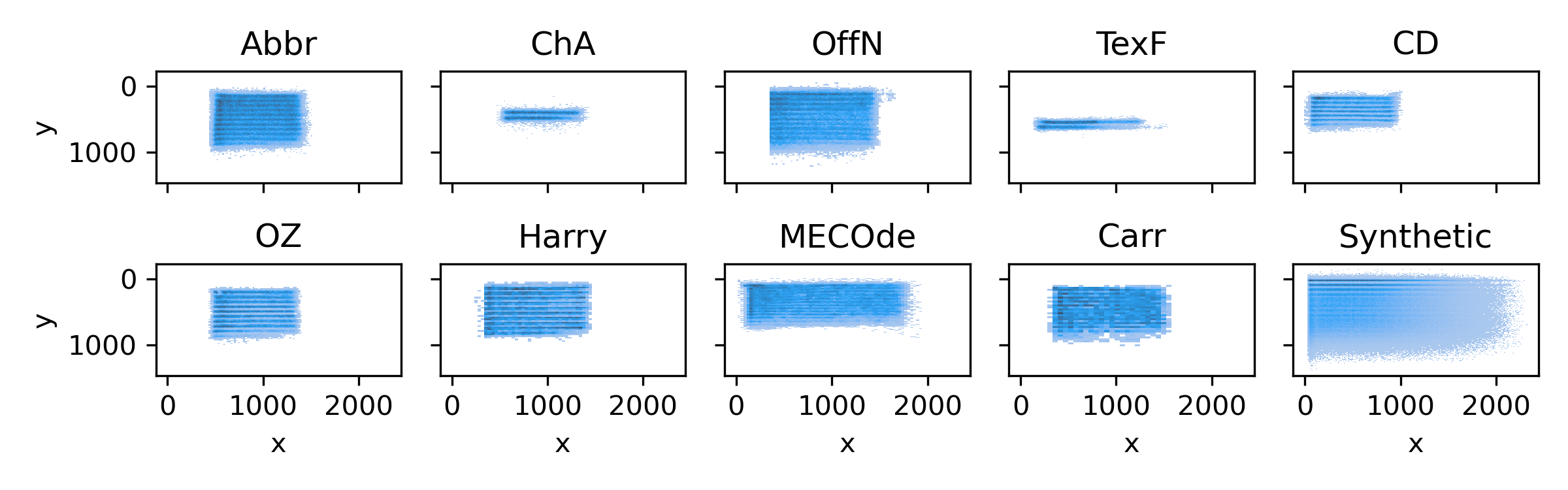}	
		\caption{No normalization with large spread of position and extend.}
		\label{fig:raw_dsets}
	\end{subfigure}
	\begin{subfigure}{0.99\linewidth}
		\centering 
		\includegraphics[width=1.0\linewidth]{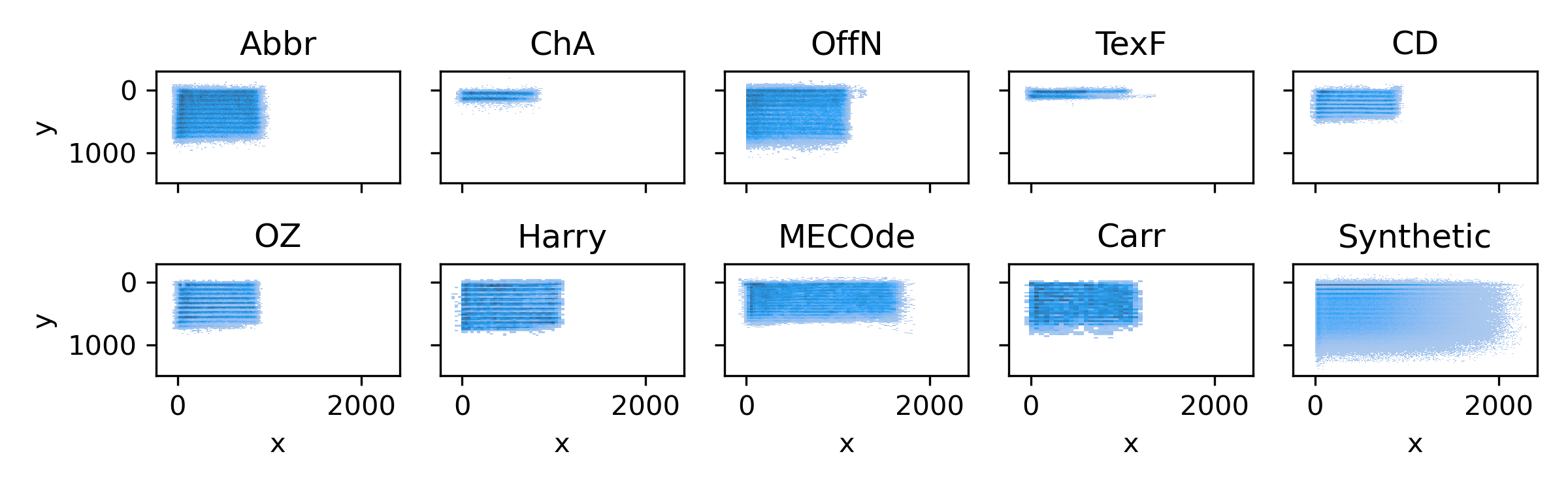}	
		\caption{\hl{After only applying xy-norm.}}
		\label{fig:xyminsubstracted_dsets}
	\end{subfigure}
	\begin{subfigure}{0.99\linewidth}
		\centering 
		\includegraphics[width=1.0\linewidth]{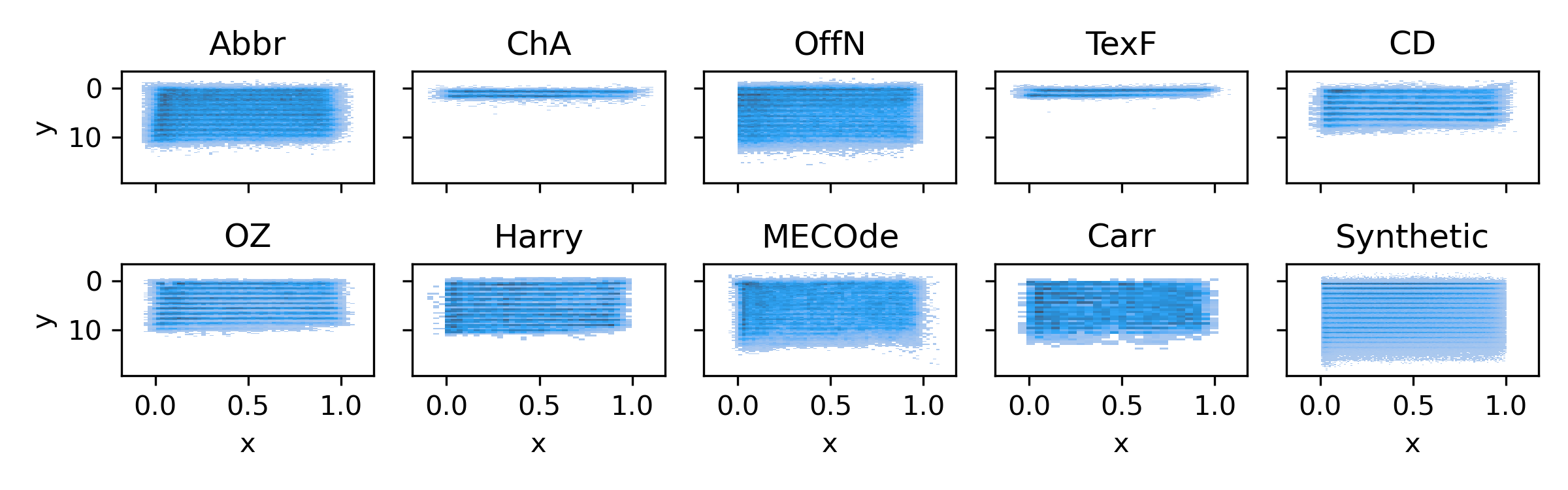}	
		\caption{\hl{After applying both xy-norm and lw-norm.}}
		\label{fig:xyminsubstractedANDheightwidth_dsets}
	\end{subfigure}
	\caption{\hl{Fixation density plot illustrating the distribution of the fixation point coordinates for all datasets before and after normalizing the fixation points by subtracting the minimum character bounding box coordinates in each trial (xy-norm) and dividing by the line width and line height (lw-norm). Darker blue colors indicate a higher concentration of fixation points across the trials. The x- and y-axes in Subfigures a) and b) give the coordinates in pixels while c) is fully normalized and thus does not have any units. Please see supplementary information for an enlarged version of this figure.}}
	\label{fig:fixations_plots}
\end{figure}

\figurename~\ref{fig:raw_dsets} shows the raw fixation point distributions for all datasets highlighting the data diversity across studies. \hl{Notably, the fixation distributions show a pattern of darker blue forming lines where the fixations cluster on the lines of text of the different stimuli but with significant variation around them. This clustering is visible due to the fact that within a specific dataset the stimulus materials for the different trials always have the lines of text in the same position.} The differences in the fixation distributions can be attributed to dissimilarities in the stimulus materials, namely the length, formatting and visual presentation of the passages read by participants, as well as the equipment employed to present the stimuli in each study and the experimental setup. This diversity presents a challenge since the model will likely only be able to correctly assign fixations to lines if the coordinates have a similar scale and distribution to what the model is trained on. As is shown in \figurename~\ref{fig:fixations_plots} and will be expanded on in Section~\ref{sec:ablation}, we address this by normalizing the fixation data using information from the stimuli of each trial of the various studies.

Before feeding the fixation coordinates into the model, the coordinates are normalized by first subtracting the minimum character bounding box coordinates \hl{(outer edge of bounding box)} found in the trial. \hl{We dub this scheme xy-norm.} The effect of this normalization step is illustrated in \figurename~\ref{fig:xyminsubstracted_dsets}. Additionally, the fixation coordinates are further normalized by dividing the \emph{y} coordinate by the minimum line height and the \emph{x} coordinate by the maximum line width found in the trial. \hl{We dub this scheme lw-norm.} This is done because the datasets had been recorded under different experimental conditions and the start position of each block of text is different for each dataset, as can be seen in \figurename~\ref{fig:raw_dsets}. The effect of the full normalization is illustrated in \figurename~\ref{fig:xyminsubstractedANDheightwidth_dsets}.

Since the different model inputs varied in their scale all values were further normalized by subtracting the mean and dividing by the standard deviation of all training data.

To evaluate the model's ability to generalize beyond its training data, we shall employ a 9-fold cross-validation scheme, where one of the nine datasets is in turn withheld from the training process, ensuring that the model is evaluated on separate, previously unseen data. 

\section{Framework}
\label{sec:framework}
\hl{Conceptually, our dual input approach consists of a sequence of fixation-related features and an image that contains information about both the fixations and the stimulus material being fed into a deep sequence model that uses this information to classify each fixation according to which line of text it belongs to. The sequence of fixation-related information makes up the first input stream and the image makes up the second input stream.}
The problem at hand can be described as a mapping \hl{$g: x_1, x_2 \mapsto y$} that jointly assigns each fixation in a sequence of length $s$ to a line index $y \in \left\{1,2,...,K \right\}$ with $K$ being the maximum line index in a trial, using fixation related features $x_1 \in \mathbb{R}^{f\times s}$ and associated trial related features $x_2 \in \mathbb{R}^{H\times W\times C}$. Here $f$ is the number of fixation related features, $s$ is the number of fixations, $H$, $W$ and $C$ are the height, width and number of channels in the input image associated with the information fed into the second input stream.

\begin{figure*}[t]
	\centering 
	\includegraphics[width=1.0\linewidth]{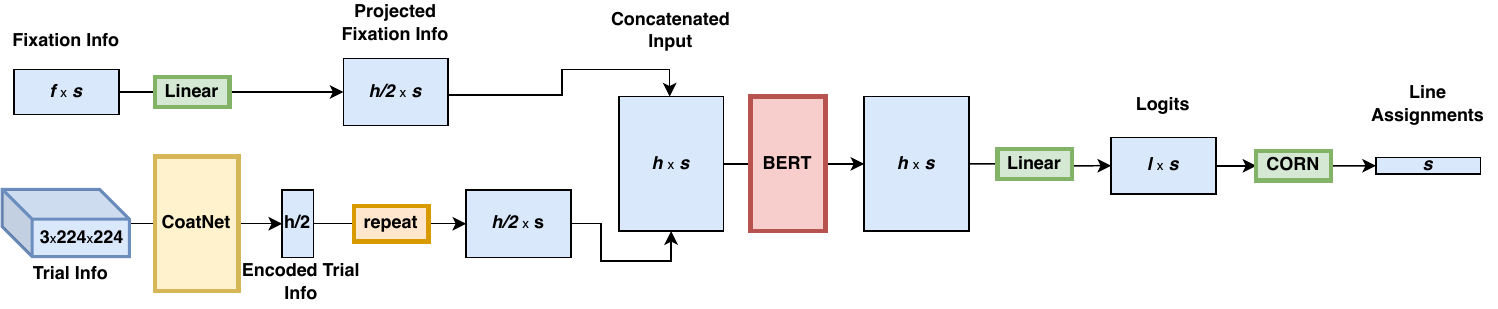}	
	\caption{Model flow with the top half showing the fixation information input stream and the bottom half showing the page information stream. \textit{\textbf{s}} is the length that all sequences are padded to. \textit{\textbf{f}} is the number of fixation related features. \textit{\textbf{h}} is the hidden dimension of the main encoder network. \textit{\textbf{l}} is the maximum number of lines in the datasets.}
	\label{fig:flow}
\end{figure*}

We leverage a bidirectional encoder-only Transformer model, \hl{while configured to be a smaller model, it} largely follows the original Bidirectional Encoder Representations from Transformers (BERT) architecture~\cite{devlinBERTPretrainingDeep2019}, as our main encoder model. \hl{Note that no pretrained weights are used for this part of our architecture. Please see \figurename~{\ref{fig:flow}} for a visual depiction of how the BERT model is positioned in the overall architecture.} The self-attention-based Transformer architecture~\cite{vaswaniAttentionAllYou2017} has proven widely successful and has largely displaced the previous recurrent network architectures due to their more direct information flow across the sequence and improved training speed and stability~\cite{zeyerComparisonTransformerLSTM2019}. These architectures largely overcome the vanishing or exploding gradient problem that recurrent neural networks suffered from when processing long sequences~\cite{goodfellowDeepLearning2016}. BERT-based approaches are able to take into account context from both directions from a specific part of the sequence~\cite{devlinBERTPretrainingDeep2019}. This makes BERT well suited to handle the long fixation sequences.

\hl{Our model incorporates two input streams: one consisting of normalized fixation features and one containing the rendered page consisting of the characters, their bounding boxes and coarsely depicted fixation point information. The fixation features consist of the \emph{x-y} coordinates of each fixation point and the line number with which the fixation point overlaps with -1 being used when a point does not overlap with any line. A fixation point is considered to overlap with a line if it's y-coordinate is within the y-coordinate-range of the character bounding boxes making up the line. No gaps exist between the bounding boxes of adjacent lines within a paragraph for all datasets except CD, which has a ten pixel gap between adjacent lines.} We use an ImageNet pretrained CoAtNet~\cite{daiCoAtNetMarryingConvolution2021,dengImageNetLargescaleHierarchical2009,rw2019timm} as a feature extractor for the second input stream. \hl{Its weights} are unfrozen, so it is allowed to train with the rest of the architecture. We choose CoAtNet due to its ability to combine the desirable inductive biases found in \hl{CNNs} with the attention mechanism. 

\begin{figure*}[t]
	\centering
	\begin{subfigure}{0.33\linewidth}
		\adjincludegraphics[width=0.85\linewidth,trim={{.025\width} {.05\width} {.11\width} {.05\width}},clip]{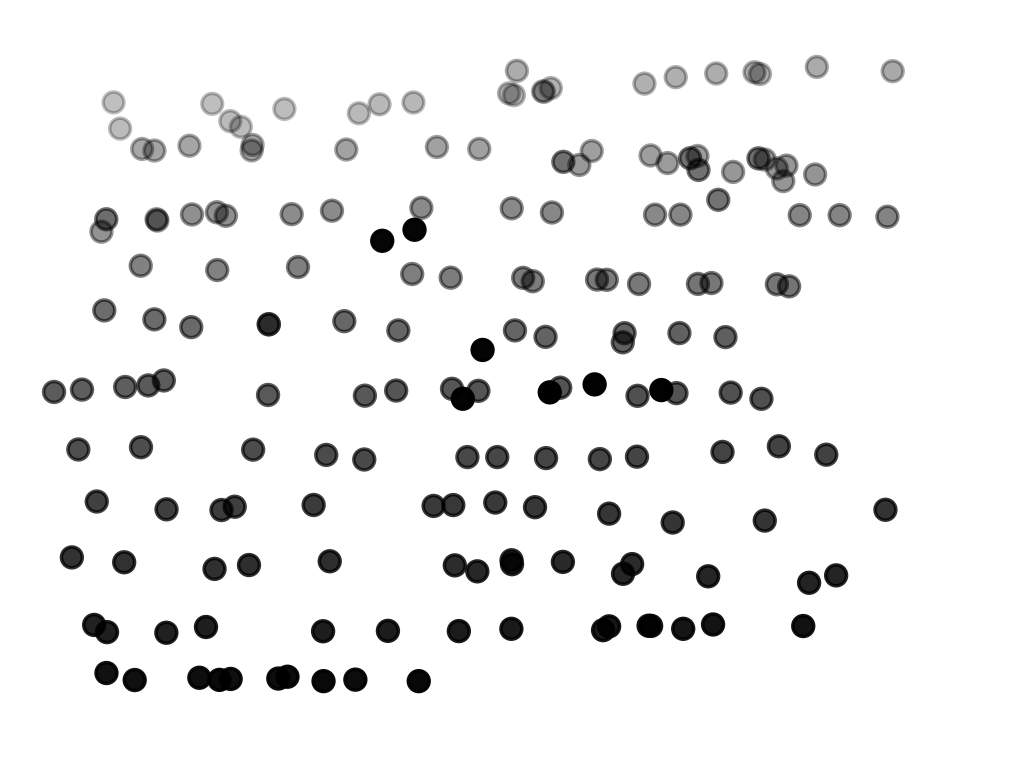}
		\caption{Fixations.}
		\label{fig:second_input_stream_fixations}
	\end{subfigure}
	\begin{subfigure}{0.33\linewidth}
		\adjincludegraphics[width=0.85\linewidth,trim={{.03\width} {.05\width} {.09\width} {.08\width}},clip]{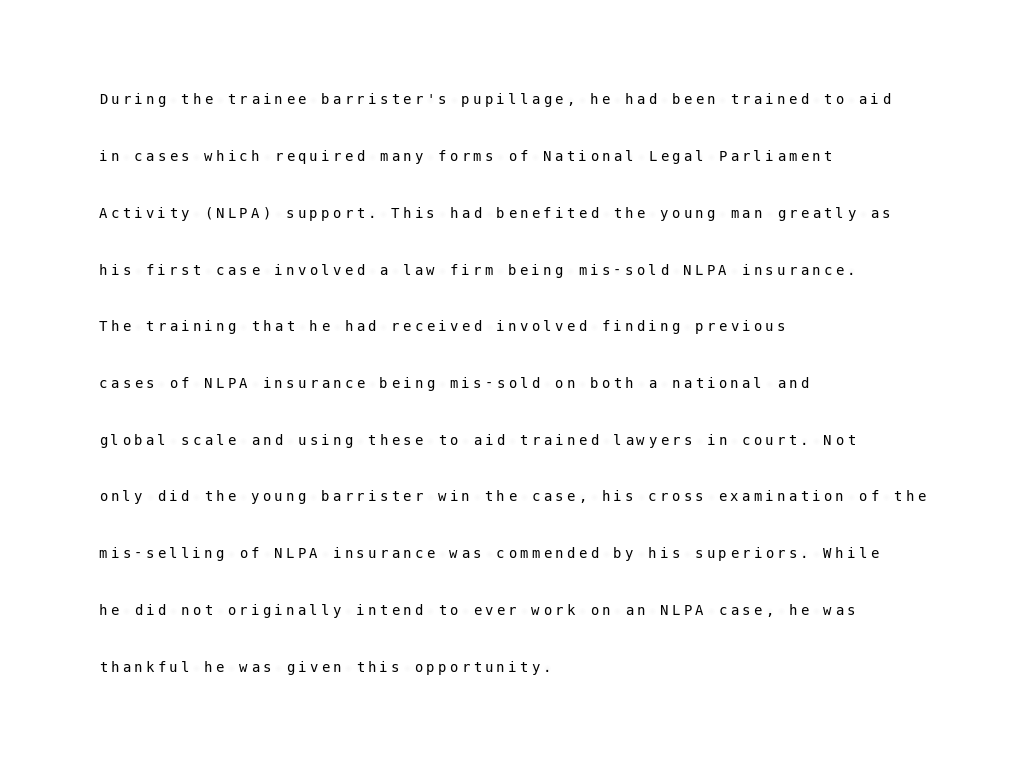}
		\caption{Characters.}
		\label{fig:second_input_stream_words}
	\end{subfigure}
	\begin{subfigure}{0.33\linewidth}
		\adjincludegraphics[width=0.85\linewidth,trim={{.025\width} {.05\width} {.11\width} {.05\width}},clip]{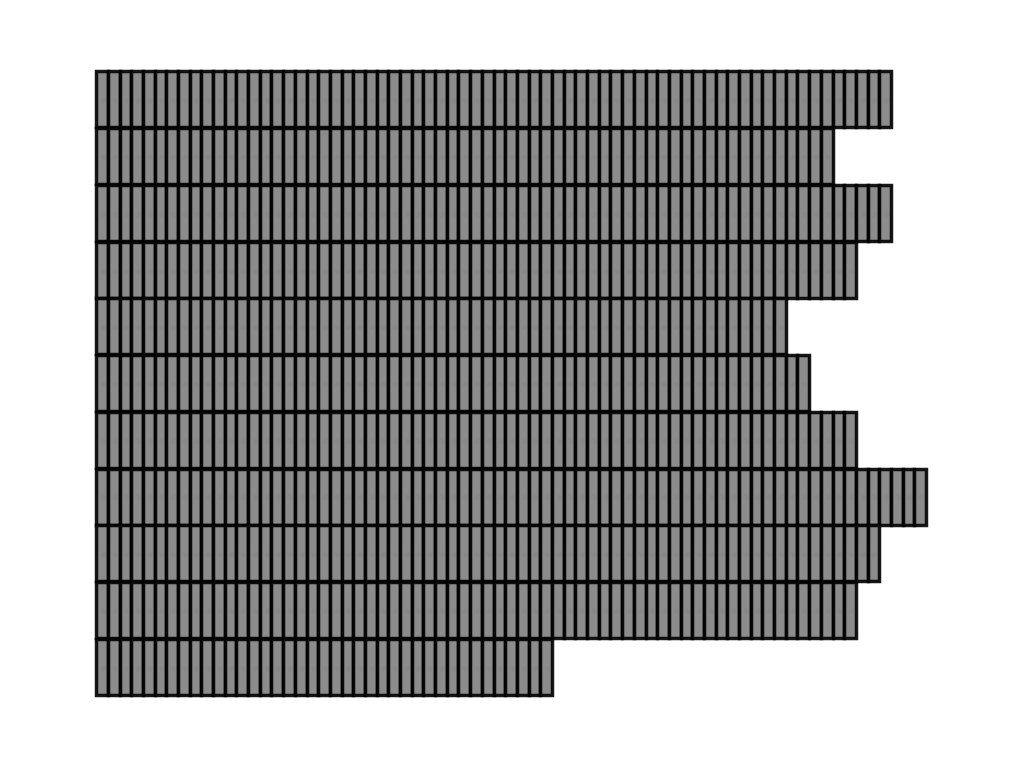}
		\caption{Bounding Boxes.}
		\label{fig:second_input_stream_boxes}
	\end{subfigure}
	\caption{Example of the single channel images used as second input stream to the main encoder.}
	\label{fig:second_input_stream}
\end{figure*}

\hl{The second input stream information is fed into the model by first grayscale rendering separate images for the stimulus text of the trial, the filled character bounding boxes and a scatter plot of the x-y-coordinates of each fixation point in the associated sequence with the fixation start time (scaled to between 0.25 and 1.0 for each trial) encoded as the gray-level of the scatter plot markers.} See \figurename~\ref{fig:second_input_stream} for a depiction of the \hl{grayscale} images used. These three single channel images are concatenated in the channel dimension to get an image with three channels as the pretrained CoAtNet expects. The CoAtNet then encodes this information into a single vector for each sequence. This vector is then projected to half the hidden dimension of the main encoder \hl{and} duplicated in the sequence dimension to match the shape of \hl{the projected fixation features. The tensors resulting from both input streams are concatenated along the feature dimension and fed into the main encoder architecture.} A simple linear head network then turns the encoder output into line prediction outputs for each fixation in the sequence. \figurename~\ref{fig:flow} shows an overview of the data flow through the architecture.

The main encoder model uses several blocks consisting of a multi-headed self-attention layer, a layer norm and a Multi Layer Perceptron (MLP). These blocks are stacked to create a deep encoder-only transformer network. The head of the network is a simple fully connected layer with no activation function. A Conditional Ordinal Regression for Neural Networks (CORN)~\cite{shiDeepNeuralNetworks2022} loss is used to train the network. It takes in the target line assignments and outputs of the last layer of the architecture, also referred to as the logits. In contrast to a simple cross-entropy loss function CORN takes the order of the line indices into account.

To feed the fixation features to the model in mini batches, each sequence of fixations has to be padded to match the longest sequence length. For the calculation of the loss and accuracy measures this padding is taken into account by masking out the padded parts of each sequence.

Since the \hl{stimulus} texts and their formatting varied across datasets, the number of possible \hl{target lines} \hl{can differ} for each trial. This is addressed by using the largest line count present across all datasets as the output dimension of the final layer. At inference time the \hl{prediction} is restricted by clipping \hl{it} to the maximum line index in the trial. It should be noted that this does not present a significant limitation of our approach since that information is always available \hl{for data from passage reading studies}.

As will be shown in Section~\ref{sec:ablation}, some datasets show very different performance depending on what kind of normalization scheme is used for the fixation data. Since it is desirable to have a single model work well for all datasets, we explore an ensemble-based approach. \hl{This is implemented by training the same DIST configuration multiple times on the same training data, randomly reinitializing all non-pretrained weights each time. This is done for each normalization scheme (normalized fixation points are illustrated in Figures \mbox{\ref{fig:xyminsubstracted_dsets}} and \mbox{\ref{fig:xyminsubstractedANDheightwidth_dsets}}).} The weights of the chosen models are then frozen and combined in an ensemble model which takes in the data with the different normalization schemes applied to it and feeds it to the list of models appropriate for each scheme. The resulting logit tensors are then averaged and fed into the CORN inference function (shown in Equation~\ref{eq:cornPred}). This requires no further model training. We refer to an ensemble of DIST models as E-DIST. 
\hl{Inspired by Carr's work\mbox{\cite{carrAlgorithmsAssigningFixations}} on combining multiple classical algorithms, we further enhance the E-DIST approach by combining it with the WOC approach, which applies multiple algorithms and uses a majority decision voting system for each fixation assignment and works as follows. For a given trial, first a set line assignments is computed for each algorithm that is to be included in the WOC approach. For each fixation in that sequence, the WOC algorithm counts how many of the algorithms assign the fixation to a line number and the line number with the highest number of votes is chosen as the result. An arbitrary number of line assignment sequences can be produced and included in the WOC voting pool. One can give additional weight to the vote of an algorithm by adding multiple copies of its line assignments to the voting pool. We refer to the approach that uses multiple copies of the line assignments produced by E-DIST as E-WOC. To distinguish this from the using only classical algorithms in the WOC voting pool, we refer to the classicals only approach as C-WOC}. We will show in Section \ref{sec:results} that \hl{E-WOC} \hl{achieves the highest accuracy of all approaches when averaged across datasets.}

\section{Implementation}
Our approach has been implemented in \textit{PyTorch}~\cite{paszkePyTorchImperativeStyle2019b}. We have configured the main encoder to have a depth of four, an internal representation dimension of 512 and eight attention heads, \hl{this is equivalent to the "Small" configuration described in \mbox{\cite{turcWellReadStudentsLearn2019}}}. The page encoding computer vision (CV) model is configured to take in images with size of 224x224 pixels and has its weights initialized from a model trained to classify images from the ImageNet dataset. For the CoAtNet we chose a model with an embedding dimension of 512. \hl{To implement the WOC approach we modify Carr's implementation\mbox{\cite{Eyekit}}.} All training was carried out on a machine equipped with an Nvidia RTX 3090 with 24 GB of memory and an Intel i7-7700K CPU with 64 GB of memory. The \textit{PyTorch} implementation \hl{of our model is publicly available.}

\section{Training Setting \hl{and Evaluation Metric}}

As touched upon in Section \ref{sec:framework} for model supervision and evaluation during training we use the CORN loss function shown in Equation~\ref{eq:cornLoss}~\cite{shiDeepNeuralNetworks2022}.

\begin{equation}
\begin{aligned}
\label{eq:cornLoss}
L(\textbf{Z},\textbf{y}) = &
- \frac{1}{\sum_{j=1}^{K-1}\left| S_j \right|}  \sum_{j=1}^{K-1}  \sum_{i=1}^{\left| S_j \right|} \left[  log(\sigma(\textbf{z}^{[i]})) \cdot \mathds{1} \left\{ y^{[i]} > r_j \right\}
\right.
\\
& 
\left.
+  (\log(\sigma(\textbf{z}^{[i]})) - \textbf{z}^{[i]}) \cdot \mathds{1} \left\{ y^{[i]} \le r_j \right\} 
\right]
\end{aligned}
\end{equation}
where $\textbf{Z}$ are the outputs of the last layer of the model, referred to as logits in \figurename~\ref{fig:flow}. $\textbf{y}$ is the set of ground truth line indices $y^{[i]}$ and $i$ the index for the model output $z^{[i]}$. $\left| S_j \right|$ denotes the size of the subset of the training data denoted by $j$, that matches the condition of the rank being no higher than $r_j$ with $r_j$ being able to take on values between 1 and $K-1$ for $K$ possible line indices. $\sigma$ is the Sigmoid function. $\mathds{1}$ is the indicator function giving 1 if the condition is met and 0 if it is not, with the condition being that the ground truth label is above or below rank $r_j$. This loss function takes into account that the line index is ranked and the difference between line numbers is meaningful by creating conditional training sets for each rank, in this case line number. For models trained using the CORN loss Equation \ref{eq:cornPred} computes the predicted line indices $q^{[i]}$. 
\begin{equation}
\label{eq:cornPred}
q^{[i]} = 1 + \sum_{j=1}^{K-1} \mathds{1}\left ( \hat{P}\left ( y^{[i]} > r_j \right ) > 0.5 \right )
\end{equation}

where $\hat{P}$ is the predicted probabilities.
CoAtNet \hl{is initialized} from pretraining on ImageNet, while weights of the rest of the architecture are randomly initialized\hl{(Kaiming initialization~\mbox{\cite{heDelvingDeepRectifiers2015}})}. We use the Adam optimizer with an initial learning rate of $4.5e-4$ and a learning rate scheduler dividing the learning rate by 2 \hl{every 3000 training steps. We train the model for 25000 training steps.}

\hl{To compare the performance of our model to that of the classical algorithms we use relative accuracy $\alpha_r$, which is the difference between our model's accuracy $\alpha_m$ and the best classical algorithm's accuracy $\alpha_c$.}

\begin{equation}
\label{eq:relAcc}
\alpha_r = \frac{\alpha_m - \alpha_c}{\alpha_c}
\end{equation}

\section{Results and Discussion}
\label{sec:results}

\begin{table*}[!ht]
	\centering

	\caption{Comparison of mean accuracy for all \hl{classical algorithms, C-WOC, DIST, E-DIST and E-WOC}. \textbf{Bold} numbers highlight the best performing approach overall and \underline{underlined} values indicate the best classical approach for each dataset. \hl{DIST accuracies are based on using input that had both xy-norm and lw-norm applied.} E-DIST accuracy is based on using \hl{three instances of the DIST model for each normalization scheme, so six instances in total}. \hl{Note six instances were chosen based on the fact that adding additional instances has diminishing returns with the OZ dataset even showing a slight drop in accuracy, as shown in the supplementary information, while increasing the computational burden. The reported accuracy for E-WOC is based on allocating three votes to E-DIST in the voting pool.}}
	\label{tbl-classicCompare}
	\begin{tabular}{ccccccccccc}
	\hline
	\textbf{Method}  & \textbf{Abbr}  & \textbf{CD}    & \textbf{Carr}  & \textbf{ChA}   & \textbf{Harry} & \textbf{MECOde} & \textbf{OZ}    & \textbf{OffN}  & \textbf{TexF}  & \textbf{Mean}  \\ \hline
	\textbf{E-WOC}   & 97.82          & \textbf{98.43} & 98.13          & \textbf{99.11} & 98.56          & \textbf{96.33}  & \textbf{98.41} & \textbf{97.00} & \textbf{99.72} & \textbf{98.17} \\
	\textbf{E-DIST}  & \textbf{98.15} & 98.22          & \textbf{99.51} & 99.09          & \textbf{98.77} & 95.26           & 98.36          & 92.01          & 99.61          & 97.67          \\
	\textbf{C-WOC}   & {\ul 94.98}    & {\ul 98.22}    & 96.27          & {\ul 99.00}    & 94.97          & {\ul 94.83}     & {\ul 98.39}    & 94.42          & {\ul 99.71}    & {\ul 96.75}    \\
	\textbf{cluster} & 94.86          & 93.88          & 94.12          & 98.06          & {\ul 98.37}    & 90.10           & 96.65          & {\ul 94.61}    & 99.61          & 95.59          \\
	\textbf{DIST}    & 97.76          & 96.00          & 98.79          & 98.95          & 98.47          & 95.31           & 97.85          & 75.18          & 99.56          & 95.32          \\
	\textbf{merge}   & 94.62          & 94.85          & 94.20          & 97.22          & 97.83          & 81.16           & 90.32          & 93.97          & 98.84          & 93.67          \\
	\textbf{regress} & 86.83          & 95.33          & 92.33          & 98.13          & 94.18          & 85.88           & 96.90          & 90.05          & 99.49          & 93.24          \\
	\textbf{slice}   & 94.15          & 93.15          & 95.70          & 98.10          & 97.21          & 94.33           & 95.22          & 71.57          & 99.59          & 93.23          \\
	\textbf{warp}    & 89.71          & 88.83          & {\ul 96.68}    & 96.25          & 88.52          & 87.69           & 94.28          & 91.90          & 95.50          & 92.15          \\
	\textbf{chain}   & 81.92          & 96.47          & 90.51          & 96.45          & 81.90          & 82.72           & 96.77          & 85.50          & 97.56          & 89.98          \\
	\textbf{stretch} & 78.74          & 90.51          & 93.97          & 97.59          & 88.30          & 82.78           & 94.67          & 83.11          & 98.35          & 89.78          \\
	\textbf{attach}  & 80.58          & 95.48          & 87.38          & 96.16          & 81.37          & 80.29           & 96.23          & 84.61          & 97.38          & 88.83          \\
	\textbf{split}   & 76.02          & 90.89          & 89.96          & 94.91          & 79.18          & 81.21           & 89.09          & 81.98          & 95.68          & 86.55          \\
	\textbf{segment} & 78.82          & 79.93          & 80.19          & 95.42          & 79.75          & 67.34           & 71.65          & 79.53          & 94.41          & 80.78          \\
	\textbf{compare} & 52.61          & 50.35          & 63.58          & 92.71          & 44.01          & 59.88           & 46.01          & 80.18          & 85.45          & 63.86          \\ \hline
\end{tabular}
\end{table*}

To provide a fair performance benchmark for our proposed \hl{approaches}, we evaluate eleven classical line assignment algorithms and C-WOC on all datasets. We keep the evaluation dataset out of the training data entirely and evaluate the model's performance on that dataset. In this manner, we perform a full cross-validation for all datasets. 

\begin{figure}[ht]
	\centering 
	\includegraphics[width=1.0\linewidth]{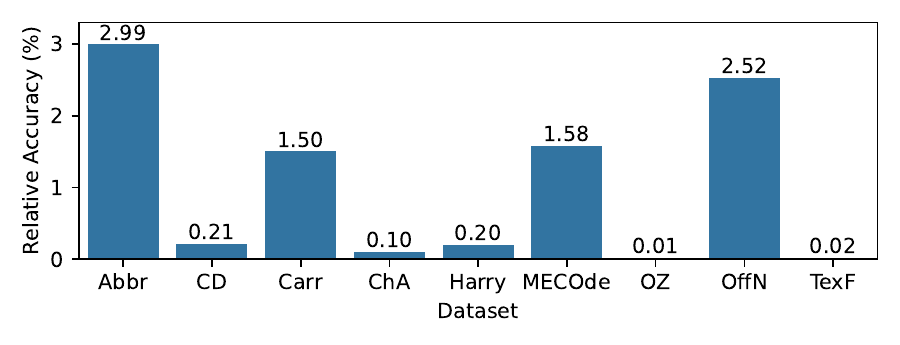}	
	\caption{\hl{Cross-validation using E-WOC evaluated using average accuracies relative to the accuracy of the best classical algorithm for each dataset. The reported accuracy is based on allocating three votes to E-DIST in the voting pool.}}
	\label{fig-crossvalbest}
\end{figure}

In Table~\ref{tbl-classicCompare} we show the average accuracy for \hl{eleven} different \hl{classical algorithms as well as their combination via C-WOC and our DIST, E-DIST and E-WOC approaches}. The accuracy metric is calculated for each trial in the evaluation dataset and then averaged across all trials in that dataset. For each dataset the accuracy of the best performing classical approach is underlined, these accuracy values were used to calculate the relative accuracies shown in \figurename~\ref{fig-crossvalbest} and used for the ablation studies in Section~\ref{sec:ablation}. \hl{The relative accuracy is the difference between our model accuracy and the accuracy of the best performing classical model (see Equation \mbox{\ref{eq:relAcc}}). This presents the most challenging comparison and could be considered somewhat unrealistic as a practitioner looking for the best way to automatically correct their fixation data would have no way of knowing which algorithm performs best for their data since they would have no ground truth data and can thus not compare the different algorithms. We nevertheless chose this measure in order to demonstrate that our proposed approach is likely to be the best default choice in most scenarios. As it can be seen, E-WOC outperforms all classical algorithms(including C-WOC) on all datasets. Note that for the TexF and ChA, the absolute accuracy achieved by C-WOC is already 99~\% or higher, since these datasets are the result of experiments using only two lines of text in their stimuli, there is little room for improvement over the classical approach. 
The strongest relative improvements achieved by E-WOC are seen for the Abbr, MECOde and OffN datasets.} It is notable that many of the classical algorithms show good performance for at least some of the datasets in question with the overall best performing classical \hl{approach being C-WOC, closely followed by \mbox{\textit{cluster}} and \textit{merge}. Furthermore, it can be seen that E-DIST achieves the highest accuracy on three datasets, emerging as the second-best method.}
 
\hl{As is shown in the supplementary information}, the performance dependence on the number of DIST instances in E-DIST is different for each dataset. The accuracy reported in Table~\ref{tbl-classicCompare} is based on using\hl{ three DIST instances for each normalization scheme, therefore six instances in total. Note that the use of six instances does not present a significant computational burden since the inference time for each instance is small and the majority of users in the field are unlikely to want to correct more than a few thousand trials at a time. Therefore, a small gain in accuracy is likely well worth the extra computation.} A single DIST model using the fully normalized fixation features as well as the second stream information as input outperforms the classical \hl{approaches on three datasets with OffN showing the lowest accuracy. OffN is the only dataset containing paragraph breaks and is thus quite different from the other datasets.}

In \figurename~\ref{fig-crossvalbest} we show the relative accuracy for each dataset achieved by \hl{E-WOC}. \hl{As it can be seen E-WOC outperforms the classical approaches on all datasets with Abbr, OffN and MECOde showing the largest gains.}

\begin{figure}[ht]
	\centering 
	\includegraphics[width=0.9\linewidth]{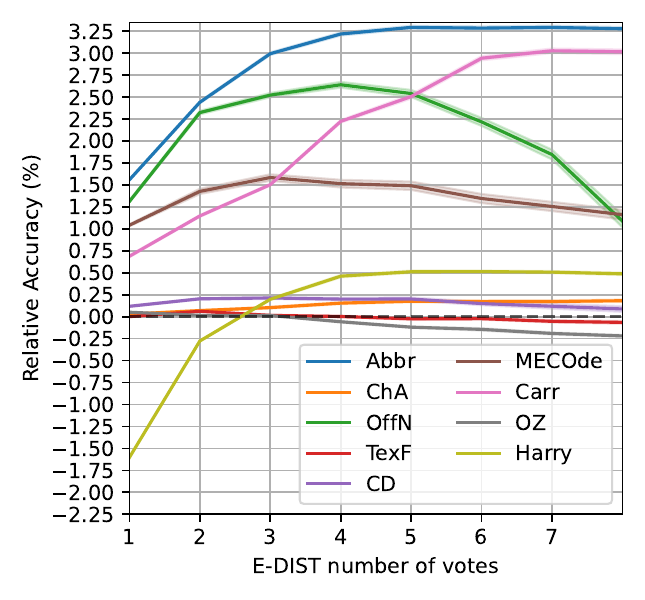}	
	\caption{\hl{Relative accuracy for E-WOC depending on the number of votes allocated to E-DIST model, which uses six DIST instances. To take into account the differences in achieved accuracy caused by which DIST instances are used in E-DIST, the data shown is the result of running the experiment 15 times for each dataset with the included instances being chosen at random (uniform probability) for each repetition.}}
	\label{fig:edist_in_woc_num_votes}
\end{figure}

\hl{In \mbox{\figurename\ref{fig:edist_in_woc_num_votes}} we show how the performance of E-WOC depends on how many votes are allocated to its predictions. Note the number of votes allocated to all other algorithms is kept at one. As it can be seen, the performance of E-WOC is strongly dependent on the number of votes allocated to E-DIST, with four of the datasets showing diminishing performance gains as the number of votes is increased. The rest of the datasets show decreasing performance for higher number of votes. As can be expected for datasets where E-DIST achieves lower accuracy than C-WOC, the performance decreases as the classical algorithms loose relative importance in the voting pool.}

\begin{figure}[ht]
	\centering 
	\includegraphics[width=0.9\linewidth]{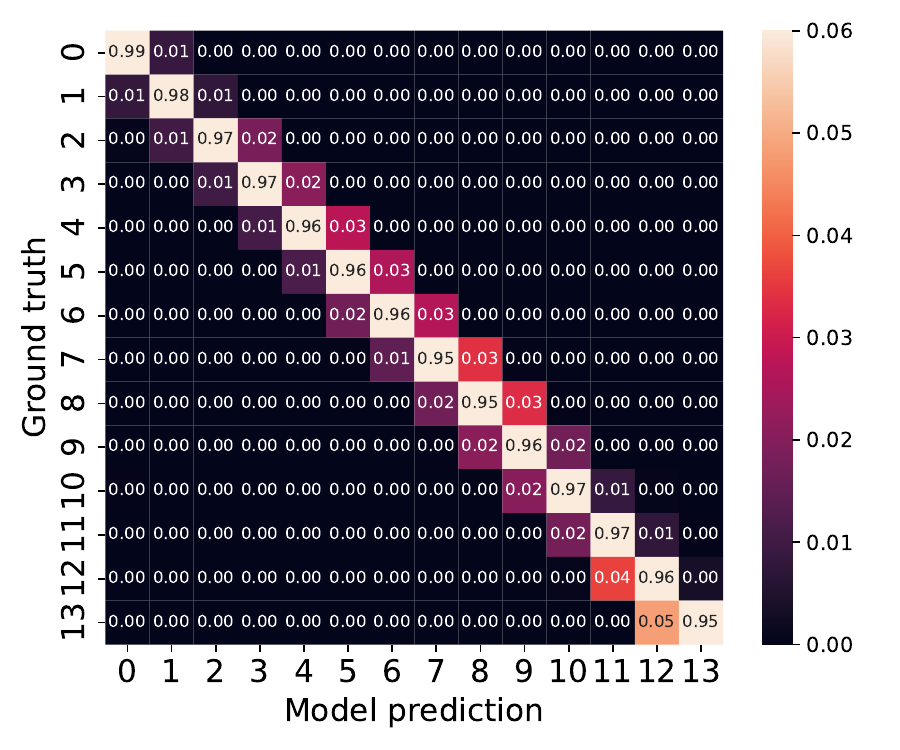}	
	\caption{Confusion matrix for all datapoints in all datasets combined. \hl{The results are based on an E-WOC with three votes allocated to E-DIST which uses six DIST instances. Note that the values are normalized for each row of the confusion matrix and the colormap is restricted to between 0 and 0.06 to better illustrate the mistakes.}}
	\label{fig:cmat_all}
\end{figure}

We show the confusion matrix for \hl{E-WOC} for all datasets in \figurename~\ref{fig:cmat_all}. As the confusion matrix shows, virtually all mistakes are misassignments by a single line. As expected from the high accuracies achieved on all datasets, the vast majority of fixation points get assigned to the correct lines.

\section{Ablation Studies}
\label{sec:ablation}

To further analyze the model's performance, we carry out a range of ablation studies. To better understand how each investigated factor of our model architecture and data pipeline affects the performance on the various datasets a full cross-validation is carried out for every ablation study. Each configuration is run at least ten times with \hl{all non-pretrained} weights of the model being randomly initialized each time\hl{(Kaiming initialization~\mbox{\cite{heDelvingDeepRectifiers2015}})}. We present the average accuracy relative to the best performing classical approach and the standard error associated with the repeated experiments. The ablation studies are carried out for a single DIST model for each experiment, rather than an ensemble of models. This is done to show the variance in the performance for the DIST model and to more clearly show the effects of each investigated factor.

\begin{figure}[ht]
	\centering 
	\includegraphics[width=1.0\linewidth]{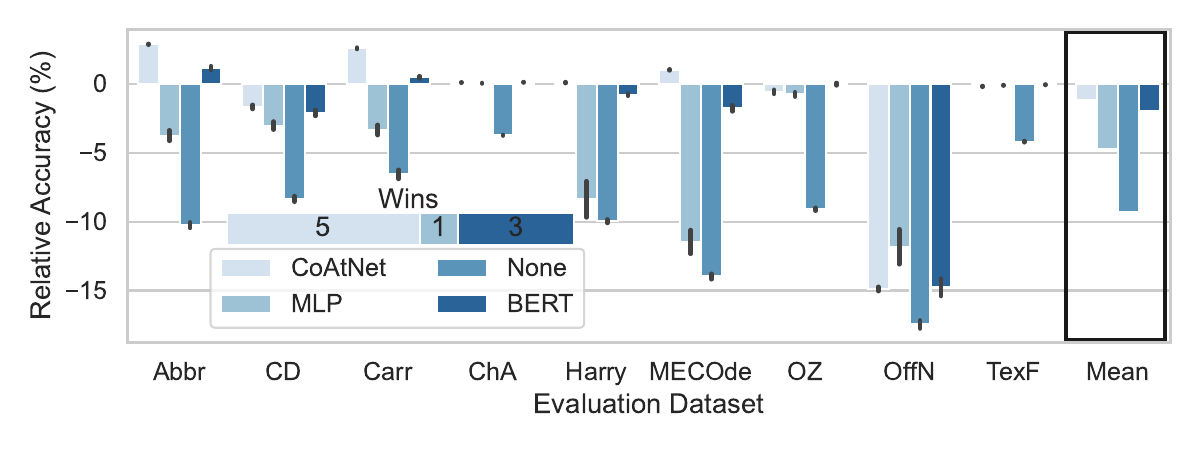}	
	\caption{Cross-validation for DIST model without the second input stream (labeled None), and for the different ways of encoding the trial related information. Respectively, BERT, MLP and CoAtNet indicate which model is used to process the trial information. \plotexplain}
	\label{fig-charenc}
\end{figure}
First, the effect of utilizing the dual input stream is investigated by training the model to predict line assignments based on the fixation information alone as well as with different methods of adding additional information to the main model input.
\hl{In addition to the CoAtNet approach described in section}~\ref{sec:framework}, we experiment with creating a sequence of bounding box coordinates of all characters on the page used as stimulus during the trial. Since this sequence length is not the same as the number of fixations in the trial, two approaches of preparing the bounding box information are explored. The first approach is to use a simple MLP consisting of two fully connected layers to first project the bounding box \emph{x-y} coordinates into a single value for each step in the sequence of characters and then passing the result to a second linear layer that projects the sequence dimension to half the embedding dimension of the main encoder. The second approach is to use a separate BERT model to encode the projected bounding box coordinates. As with the CoAtNet-based approach, this results in a single encoding vector for one sequence of fixations, which then gets repeated for every step in the fixation sequence. For all methods of including the second input stream the resulting tensor is concatenated with the projected fixation related information from the first input stream and fed into the main encoder.
In \figurename~\ref{fig-charenc} we show that the model greatly benefits from using a dual input stream approach with the accuracy dropping strongly for all datasets when only the fixation related input stream is used. \hl{DIST is outperformed by the classical algorithms for all datasets when no 2nd input stream is utilized. The MLP encoding approach only outperforms the classical approaches for one dataset. The BERT character bounding boxes encoding method outperforms the classical algorithm for three. Overall, the CV-based approach greatly outperforms all other approaches, achieving the highest accuracy on five datasets.} The requirement of a second input stream does not present a restriction of the model since most classical models also rely on page related information to correct fixations and this information is \hl{typically} recorded during the eye-tracking experiment and thus readily available.

\begin{figure}[ht]
	\centering 
	\includegraphics[width=1.0\linewidth]{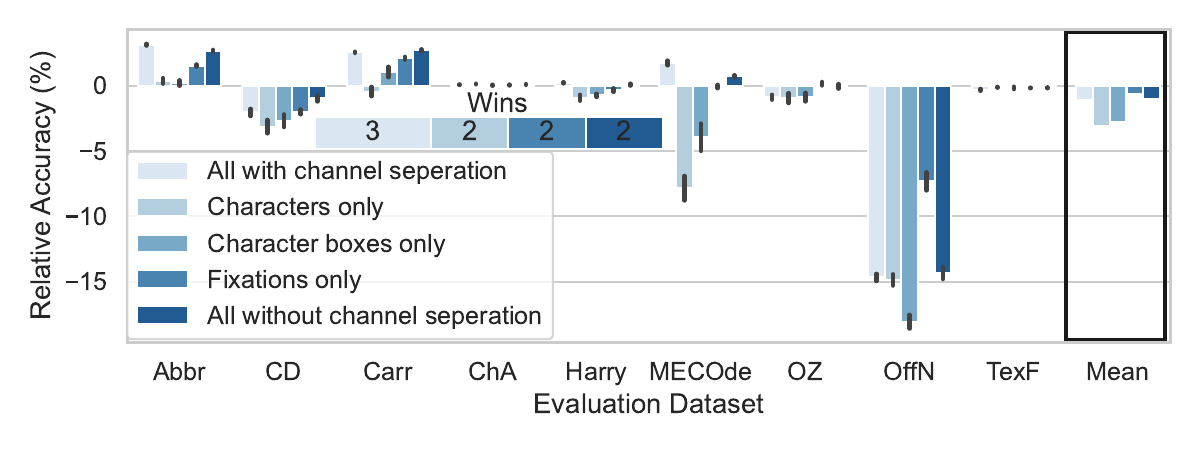}	
	\caption{\hl{DIST} performance for using different input images as the second input stream as input to the CoAtNet.}
	\label{fig:charcvinput}
\end{figure}

In \figurename~\ref{fig:charcvinput} we show how the DIST model performs when different kinds of page and fixation related information are fed into the model as a second input stream. Since the pretrained CoAtNet model that is used to encode this information expects an input with three channels, the single channel input rendering is duplicated three times in the channel dimension before being fed into the model. As it can be seen, using a scatter plot of the fixation positions without any text or bounding box information shows much better performance than using only the characters or character bounding box images for the OffN dataset. The magnitude of this difference pushes up the average performance for this input type. However, for most of the datasets the combination of all three types of information results in the best performance. The concatenation of all three renderings in the channel dimension and the non-concatenated color rendering give roughly equal average accuracy but the former achieving the highest accuracy on more datasets. OffN \hl{is} a particularly challenging dataset, as it is the only one with paragraph gaps in the stimulus material, resulting in a very different pattern of fixation sequences.

\begin{figure}[ht]
	\centering 
	\includegraphics[width=1.0\linewidth]{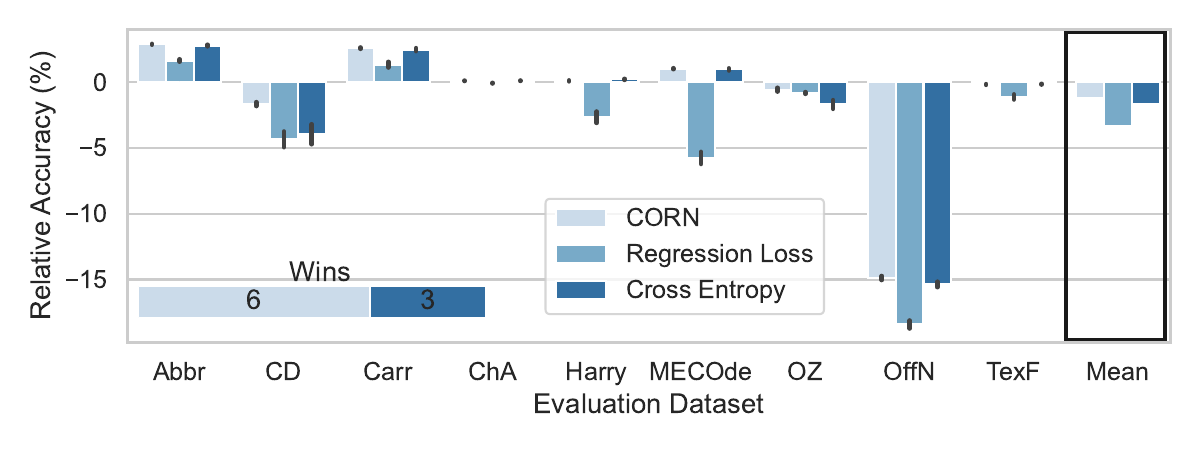}	
	\caption{Cross-validation for DIST with training done by using different loss functions to supervise.}
	\label{fig-lossfunc}
\end{figure}

\figurename~\ref{fig-lossfunc} shows how the model performs depending on which loss function is used to supervise the model during the training process. The rank consistent CORN loss produces the best overall performance as judged by the number of datasets on which it leads to the highest accuracy, closely followed by a standard cross-entropy loss, which does not take the ordering information into account. Both the CORN and the cross-entropy loss functions achieve similar average accuracies across the datasets. The regression loss, which is a simple mean squared error for treating the line assignment as a continuous number between zero and the maximum line in all datasets, does poorly on most datasets, showing the lowest average accuracy and no wins.

\begin{figure}[ht]	
	\centering
	\begin{subfigure}{0.99\linewidth}
		\includegraphics[width=0.99\linewidth]{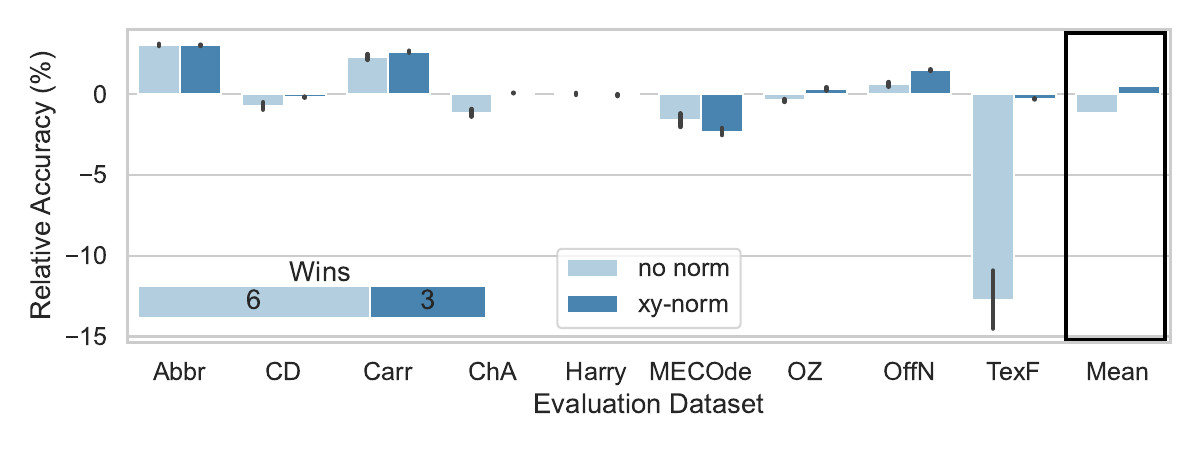}
		\caption{\hl{Comparison of performance for using only xy-norm}.}
		\label{fig:normalization1}
	\end{subfigure}
	\begin{subfigure}{0.99\linewidth}
		\includegraphics[width=0.99\linewidth]{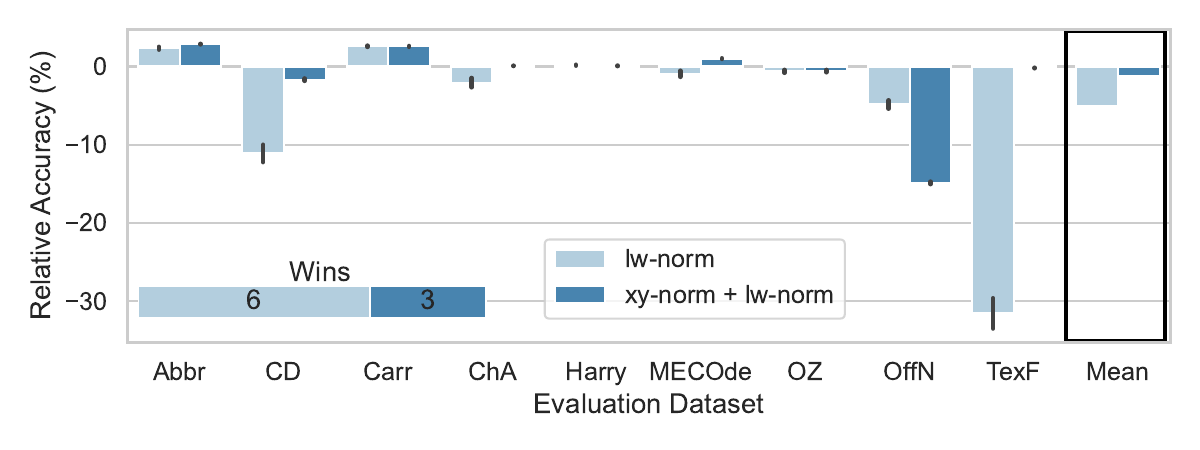}
		\caption{Effect of \hl{using only lw-norm compared to using both xy-norm and lw-norm}.}
		\label{fig:normalization2}
	\end{subfigure}
	\begin{subfigure}{0.99\linewidth}
		\includegraphics[width=1.0\linewidth]{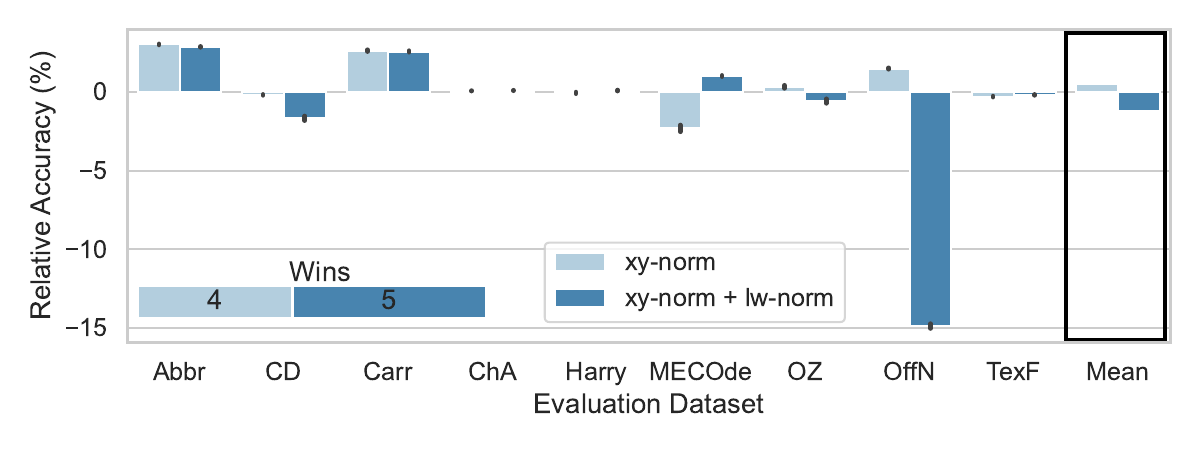}	
		\caption{Effect of \hl{using only xy-norm compared to using both xy-norm and lw-norm}.}
		\label{fig:normalization3}
	\end{subfigure}
	\caption{Effect on relative accuracy of using different normalization schemes for the fixation points of the first input stream. \hl{Note that when both schemes are applied, the xy-norm is always applied first.}}
	\label{fig-normalization_ablation}
\end{figure}

\figurename~\ref{fig-normalization_ablation} shows the effect of using different normalization schemes for the fixation coordinates on the DIST model performance on the various datasets. As it can be seen, the chosen normalization scheme has a major effect on model performance for some datasets with the CD, OffN and MECOde seeing the biggest effect. \figurename~\ref{fig:normalization1} shows that if \hl{lw-norm is not used, the effect of using xy-norm} is small for \hl{five} datasets, while retaining an overall positive effect on the mean of all relative accuracies.

\figurename~\ref{fig:normalization2} shows the effect of \hl{using lw-norm with and without xy-norm}. It can be seen that most datasets benefit from this combination while the OffN dataset sees a strong drop in relative accuracy.

\figurename~\ref{fig:normalization3} shows the effect of \hl{using xy-norm with and without lw-norm}. As it can be seen, the difference in performance is very pronounced for the MECOde, \hl{CD} and OffN datasets. The MECOde dataset performs very poorly without this normalization step while \hl{both the OffN and CD} datasets see a strong boost when the line height and width normalization is not used.

The above analysis shows that none of the schemes or its combinations is a clear best choice for all datasets. This motivates the utilization of an ensemble-based approach which can make use of the benefits of both approaches and give good performance on all datasets \hl{when used in E-WOC} as can be seen in \figurename~\ref{fig-crossvalbest}. See Section~\ref{sec:framework} for an explanation of how the different normalization schemes are used in the E-DIST model \hl{and how this is used as part of E-WOC}.

\begin{figure}[ht]
	\centering 
	\includegraphics[width=0.9\linewidth]{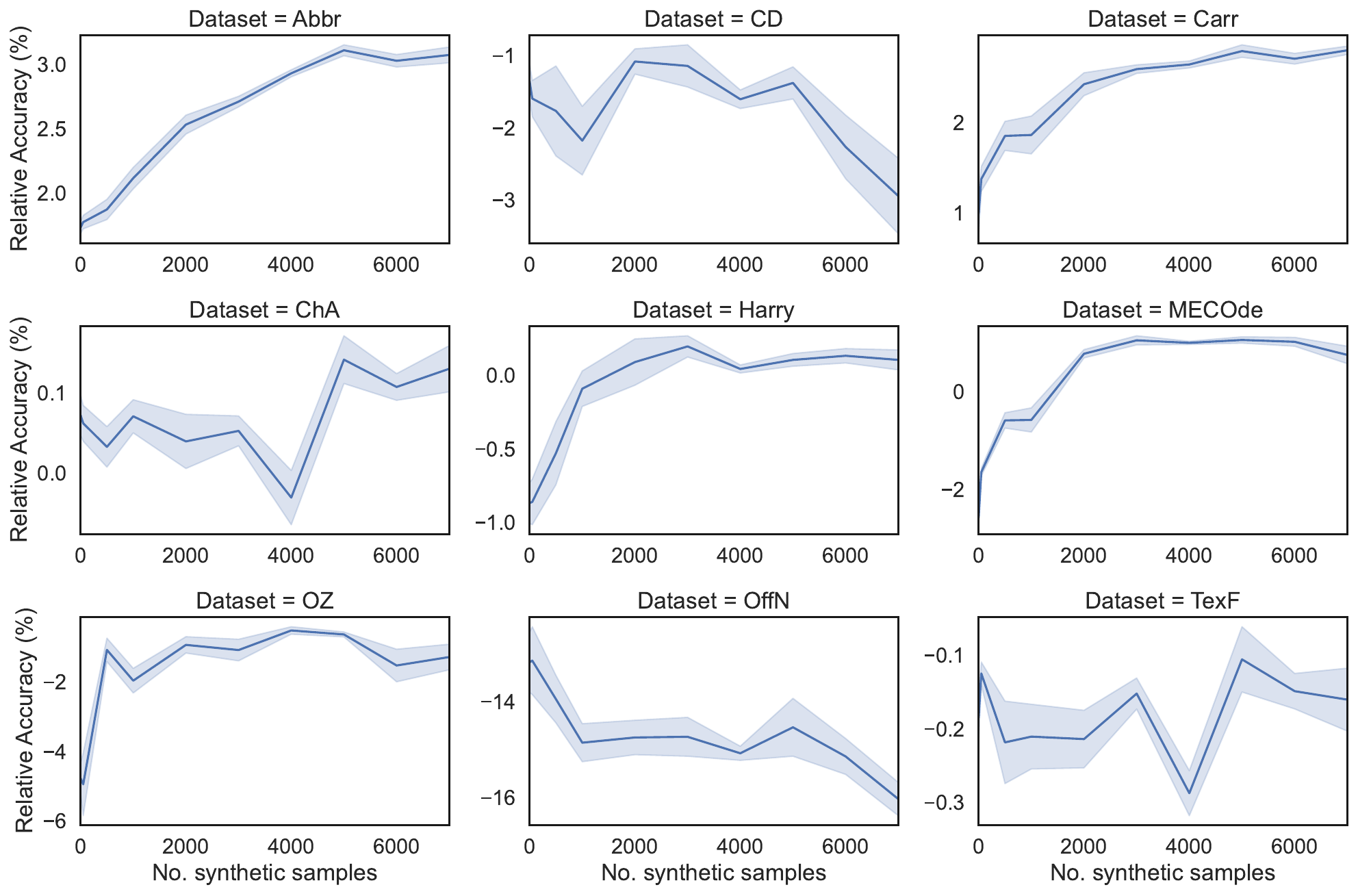}	
	\caption{Effect of using different amounts of synthetic data during training on the relative accuracy. The thick blue line indicates the mean of repeated experiments while the shaded area indicates the standard error. Note, the scale of the relative accuracy axis is not kept the same across the subplots to show the trend for each dataset.}
	\label{fig:simdataamount}
\end{figure}

In \figurename~\ref{fig:simdataamount} shows how the relative accuracy develops depending on the number of synthetically created trials in the training data. \hl{Please see Section}~\ref{sec:dsets} \hl{and the supplementary material for an explanation on how these trials can be generated.} As it can be seen, the Abbr, Carr, Harry and MECOde datasets benefit significantly from adding at least 2000 synthetic trials\hl{ to the training data}. OffN is the only dataset where adding \hl{any} synthetic trials\hl{ to the training data} results in a significant drop in relative accuracy while the accuracy of the remaining datasets is largely unaffected. As it can be seen in Table~\ref{tbl-datasets} in Section~\ref{sec:dsets} MECOde is the only dataset with 14 line long stimulus texts, therefore adding synthetically created trials with the same number of lines will likely help the model learn such assignments.

\begin{figure}[ht]
	\centering 
	\includegraphics[width=1.0\linewidth]{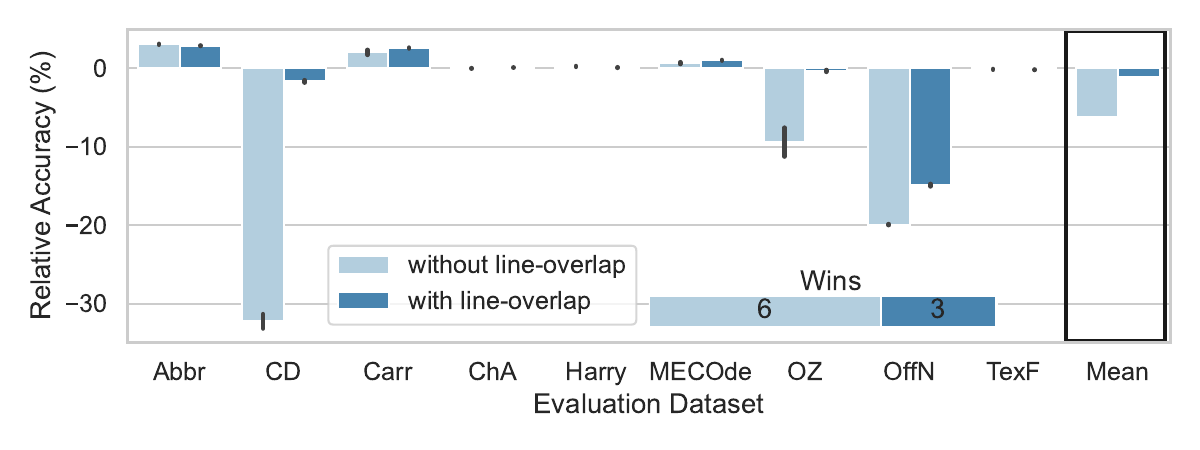}	
	\caption{Effect of adding the line-overlap feature as additional fixation related information to the first input stream.}
	\label{fig-lineoverlap}
\end{figure}
\figurename \ref{fig-lineoverlap} shows how DIST performs with and without the line-overlap feature in the first input stream. \hl{While for most datasets the addition of the line overlap feature does not change the relative accuracy by much, the CD, OZ and OffN see a large increase in performance. This results in an overall positive effect of adding the overlap feature.} Since the overall amount of training data is limited, the model clearly benefits from being given this additional information.

\section{Conclusions}

We have introduced DIST, a dual input stream architecture using BERT as its main encoder to assign fixations from eye-tracking data collected during passage reading to their most appropriate line of text. The dynamically changing vertical drift of the recorded gaze positions makes this line assignment a crucial processing step. Our architecture utilizes fixation related information as well as trial related information to map each fixation to a line index. We evaluate our model as well as \hl{eleven} classical approaches found in literature on data from nine eye-tracking studies. \hl{It is demonstrated that an ensemble of DIST models using fixation data normalized in two different ways combined with the classical algorithms in a WOC approach outperforms all classical approaches on all datasets}. The high performance of the ensemble approach is attributed to the finding that for some datasets a single DIST model performs very differently depending on how the fixation coordinates are normalized. In addition, we show that while many classical algorithms show impressive performance, the achieved accuracy \hl{varies} greatly, depending on the dataset. Our \hl{combined approach}, in contrast shows robust performance on all datasets and is hence a safe choice for any fixation alignment dataset. The addition of a line overlap feature for each fixation point emerges as crucial for achieving high accuracy for \hl{some} of the datasets. It is also demonstrated that the use of a second input stream in addition to the fixation related features greatly benefits the model performance with a CV-based approach emerging as particularly effective. We furthermore show that the inclusion of synthetic data in the training dataset is beneficial for \hl{some} datasets. Overall, our \hl{approach presents a considerable} improvement over previously published classical approaches in terms of accuracy and robustness to differences in stimulus settings. \hl{This} makes the presented method a safe first choice for practitioners, enabling them to carry out and analyze eye-tracking studies with larger amounts of text without being limited by the bottleneck of human corrected line assignments.


%

%
%
%
%
%
%

\ifCLASSOPTIONcaptionsoff
  \newpage
\fi



%

\bibliographystyle{IEEEtran} 
\bibliography{Eye-Tracking.bib}

%
%
%

%

\begin{IEEEbiography}[{\includegraphics[width=1in,height=1.25in,clip=True,keepaspectratio=True]{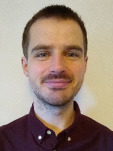}}]{Thomas M. Mercier} completed his B.Sc. in Nanoegineering at the University of Duisburg-Essen in Germany, which was followed by completing an M.Sc. and PhD in Nanoelectronics at the University of Southampton in the UK. He is currently employed as a post-doctoral researcher working on modelling of eye-movements during reading using Machine Learning techniques.
\end{IEEEbiography}

\begin{IEEEbiography}[{\includegraphics[width=1in,height=1.25in,clip,keepaspectratio]{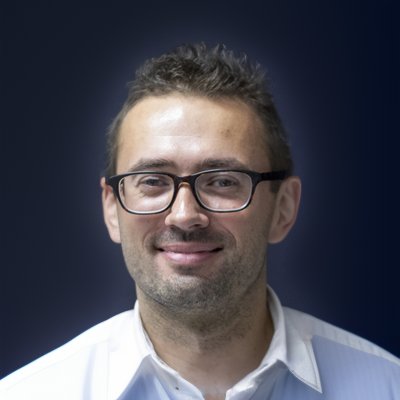}}]{Marcin Budka} received his dual M.Sc./B.Sc. degree in Finance from the Katowice University of Economics (2003), B.Sc. in Computer Science from the University of Silesia (2005) and PhD in Computational Intelligence from Bournemouth University (2010). Between 2003 and 2007 he was working as an engineer, project manager and team leader in a smart-metering start-up, before pursuing an academic career. In the years 2011-2012 he was appointed as a Visiting Research Fellow at the Wrocław University of Technology. Between 2015 and 2017 he was the Head of Research in the Department of Computing and Informatics at Bournemouth University.
\end{IEEEbiography}

\begin{IEEEbiography}[{\includegraphics[width=1in,height=1.25in,clip,keepaspectratio]{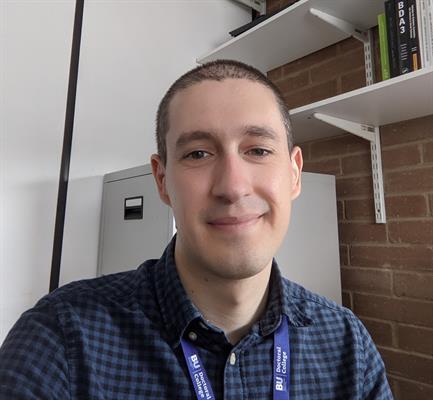}}]{Martin R. Vasilev} obtained a BA degree in Psychology from the University of Sofia (Bulgaria) in 2013 and an M.Sc. degree in Clinical Linguistics from the University of Potsdam (Germany) in 2015. He completed his PhD degree at Bournemouth University between 2015 - 2018, focusing on eye-movement control and auditory distraction. He has been working as a lecturer since 2020.

\end{IEEEbiography}

\begin{IEEEbiography}[{\includegraphics[width=1in,height=1.25in,clip,keepaspectratio]{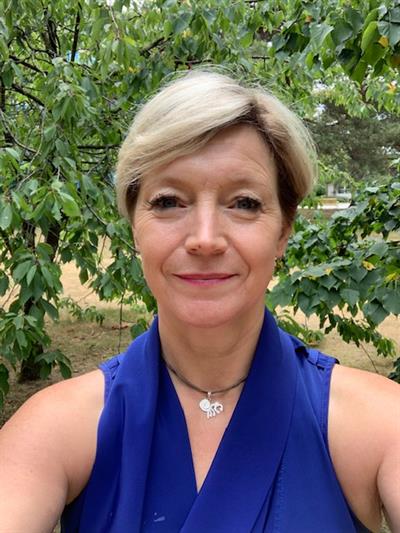}}]{Julie A. Kirkby}'s research interests fall within the field of cognitive psychology, in particular eye movements, reading and visual cognition. The primary goal of her research is to increase understanding of the causes and outcomes of developmental dyslexia. Julie earned her PhD in Cognitive Psychology under the supervision of Prof. Simon P Liversedge at the University of Southampton examining dyslexia. Since 2010, she has worked at Bournemouth University and established the Reading Research group.
\end{IEEEbiography}

\begin{IEEEbiography}[{\includegraphics[width=1in,height=1.25in,clip,keepaspectratio]{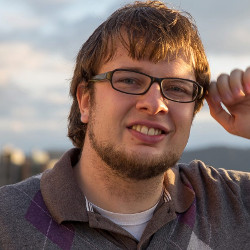}}]{Bernhard Angele} obtained an MA degree (in 2009) and a PhD degree on parafoveal processing in reading (in 2013) from the University of California San Diego. His research interests primarily focus on eye movements during skilled adult reading and language processing. Specifically, he has been studying the effect of parafoveal preview on processing and reading performance.
\end{IEEEbiography}

\begin{IEEEbiography}[{\includegraphics[width=1in,height=1.25in,clip,keepaspectratio]{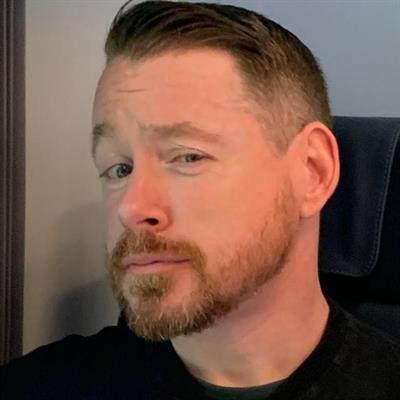}}]{Timothy J. Slattery} obtained a B.Sc. in Psychology with honors from Buffalo University in 1996 before working at a NY residential treatment center until pursuing PhD studies in Cognitive Psychology at the University of Massachusetts, Amherst in 2001, with minors in Quantitative Analysis. Post-PhD, he undertook a Post-Doc position at UC San Diego with Keith Rayner before becoming an Assistant Professor at the University of South Alabama in 2011, where he established a new Psycholinguistics lab. He joined Bournemouth University's Psychology Department in 2015 as part of their cognitive and neuroscience research team.
\end{IEEEbiography}




\end{document}


\title{Supplementary information for Dual input stream transformer for vertical drift correction in eye-tracking reading data}
\maketitle

\section{Data example}
\label{sec:dataex}

\begin{figure}[ht]
    \centering
    \includegraphics[width=0.6\linewidth]{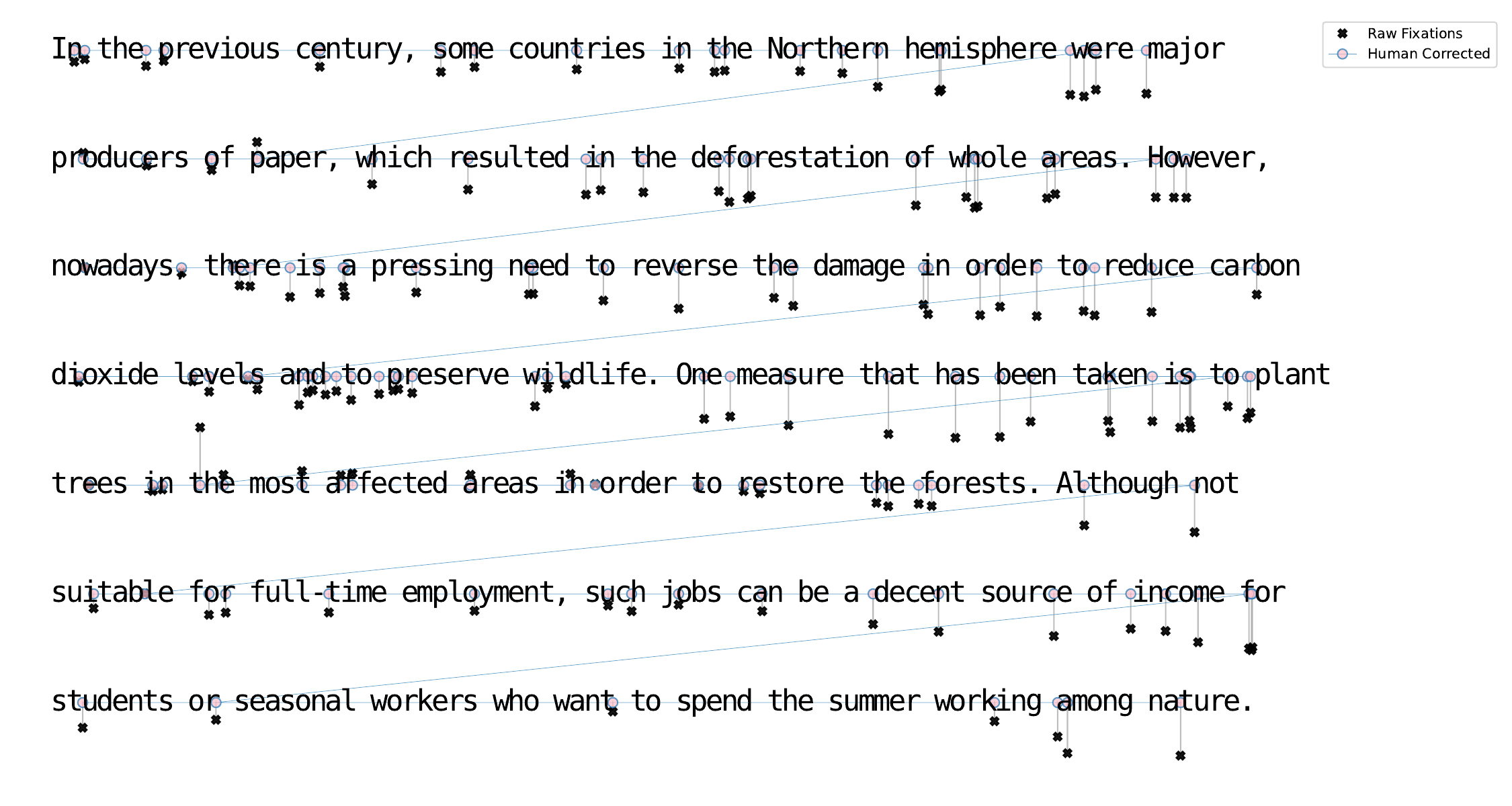}
    \caption{An example illustrating corrected and uncorrected fixations, taken from the CD dataset.}
    \label{fig:exfixplot}
\end{figure}
Fig. \ref{fig:exfixplot} illustrates the difference between corrected and uncorrected fixations for a trial from the CD dataset. As it can be seen, many fixations have drifted far below the line that the reader was actually focusing on.

\section{Data format and preprocessing}
\label{sec-dataformat}
The data for seven of the studies were recorded using the EyeLink 1000 tower setup with the eye-tracker being manufactured by SR Research (Toronto, Canada). Here the data consists of large text files containing both all the eye-tracking gaze points as well as the identified fixations. Furthermore, the files contain the bounding box information for each interest area in the stimulus text used for the trial. The interest areas are either the individual characters on the page or the words making up the passage. For these studies, the ground truth line assignments consist of output files in the plain text format typically resulting from using the Eye Doctor software developed at the University of Massachusetts Amherst~\cite{stracuzziEyeDoctorSoftware2004} to carry out the manual line assignments. For the MECOde dataset the original data consists of two .csv files, one containing the fixation coordinates, timestamps and human corrected line assignments, the other containing the character bounding box information for the stimulus texts for each trial.

All data is preprocessed to be in the same format. For each trial in each dataset a .csv file is created containing the fixation coordinates and human corrected line assignments. In addition, a .json file is created for each trial containing information about the trial and the stimulus used for the trial.

\section{Synthetic data creation}
\label{sec-synth_data_suppl}
To expand on the description given in the dataset section of the main paper this section will give more detail on the synthetic data creation. The code was adapted from Carr et al.~\cite{carrAlgorithmsAutomatedCorrection2022}. The adaptation can be found here: \href{https://github.com/Gittingthehubbing/eye_track_vdrift_algos/blob/1ec39a6bf8a11bdf48e96dd7cb0cb75618a492f3/code/simulation.py}{Code on Github}.

The code allows for the simulation of passage reading scenarios where the number of lines, number of characters per line, line spacing and level and kinds of distortion can be set.

The distortion factors used to produce the synthetic fixations are: 
Noise: Normally distributed random displacement of a fixation y-coordinate, with a standard deviation $d_{{{\mathrm{noise}}}}$ ranging from of 0 to 40
Shift: Progressive displacement of the y-coordinate of each fixation based on the y-coordinate of the center of the line that the fixation is associated with. The shift value $d_{\mathrm{shift}}$ is varied between -0.2 and 0.2.
Within Line Regression: Creation of an additional fixation that jumps back to a previous point on the current line. The x-coordinate of the regression follows a triangular probability distribution. A probability parameter determines the likelihood of the simulated reader jumping back to a previous point on the current line. A value of 1 means a regression after every normal fixation, doubling fixations on the line, while 0 implies no within-line regression. The x position of the additional fixation is randomly chosen between the line start and the current fixation, with longer regressions less likely than shorter ones. The y value follows Equation \ref{eq:distortion_suppl}.
Between line regression: Creation of an additional fixation that jumps back to a previous point on a line above the current line. The line number and the x-coordinate of the regression follow a triangular probability distribution with the peak and boundary set to the current line number and x-coordinate, respectively. For this type of regression, another probability parameter determines the likelihood of the reader going back to a previous line during the current line's reading. A value of 1 means a regression for every line read, while 0 means no between-line regression. The previous line and its section are randomly chosen, with more recent lines and sections being more probable. The y values of the regressed fixations again follow Equation \ref{eq:distortion_suppl}.

\begin{equation}
y=\mathcal{N}(l_{y},d_{{{\mathrm{noise}}}})+l_{y}d_{\mathrm{shift}}    
\label{eq:distortion_suppl}
\end{equation} 

Equation \ref{eq:distortion_suppl} shows how the y-coordinate for each fixation is calculated. Here, $\mathcal{N}$ is the normal distribution, $l_y$ is the y-center of the line, $d_{{{\mathrm{noise}}}}$ is the standard deviation of noise and $d_{\mathrm{shift}}$ is the y-shift in pixels. Note that the y-coordinate of a regression is again determined by  Equation \ref{eq:distortion_suppl} with respect to the line in question.

Please note for all randomization a uniform distribution is used unless mentioned otherwise.
For a more detailed description of the different distortion factors please see Carr et al.~\cite{carrAlgorithmsAutomatedCorrection2022}.

Based on the \emph{wikitext} dataset~\mbox{\cite{merityPointerSentinelMixture2016}} we generate

To simulate passage reading scenarios, first passages of text are created by loading in the wikitext dataset~\cite{merityPointerSentinelMixture2016} and cleaning up the texts by removing various non ASCII characters and unnecessary spaces. A minimum and maximum number of lines per passage and characters per line are then set. Generated passages consist of 8 to 14 lines of English text with each line being up to 130 characters long and the line height varying between 49 and 79 pixels. The choice for each parameter for each passage was based on a uniform random distribution. We loop over the distortion factors their range of severity values to generate fixation sequences with distortion applied to them. For each passage a sequence of up to 500 fixations is generated.

The resulting simulated trials are then saved in the same format as the real data.

The adapted code differs from Carr et al.~\cite{carrAlgorithmsAutomatedCorrection2022} in that it uses English text and how the applied noise is chosen for each trial. How the noise is applied is kept the same.

\section{Additional Results}

\subsection{Additional Ablation Studies}

\begin{figure}[ht!]
	\includegraphics[width=1.0\linewidth]{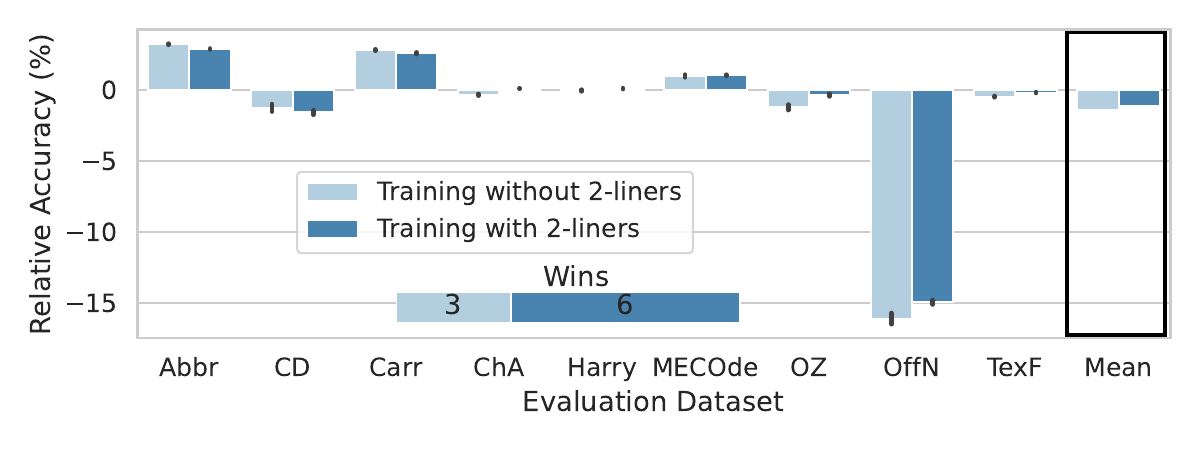}
	\caption{Training with and without 2 line datasets.}
	\label{fig:cross_val_results_2linersInTrainset_barStrip_se}
\end{figure}

In Fig. \ref{fig:cross_val_results_2linersInTrainset_barStrip_se} we show that the performance of the DIST model benefits from having the 2-line datasets inside the training data for 6 out of the 9 datasets.

\begin{figure}[ht!]
	\includegraphics[width=1.0\linewidth]{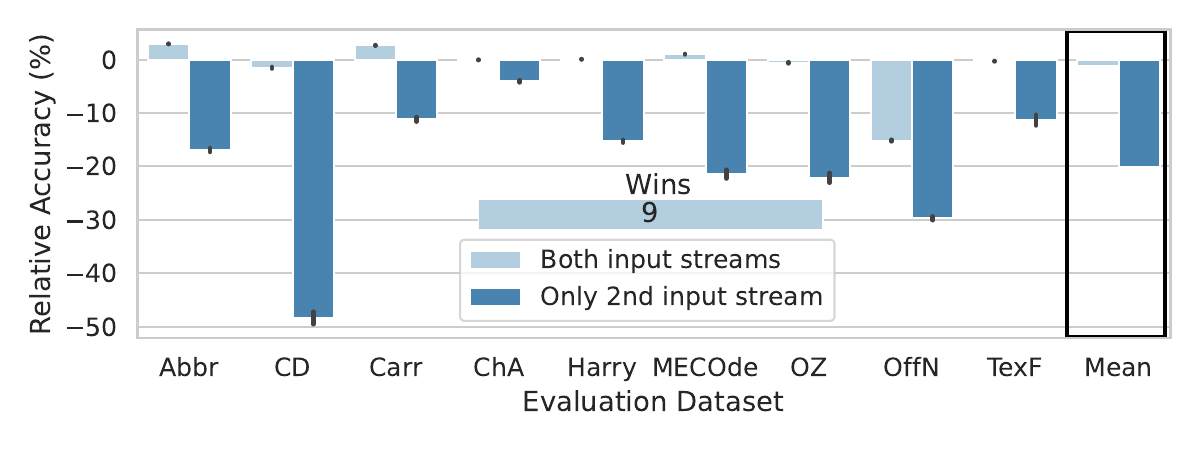}
	\caption{Training with and without using the first input stream.}
	\label{fig:cross_val_results_only_use_2nd_input_stream_bar_se}
\end{figure}
In Fig. \ref{fig:cross_val_results_only_use_2nd_input_stream_bar_se} we show how a DIST model that only uses the second input stream, which consists of an image containing both fixation and stimulus information as described in the framework section of the main paper. As it can be seen the performance drops sharply without the first input stream.

\begin{figure}[ht]
	\centering 
	\includegraphics[width=\linewidth]{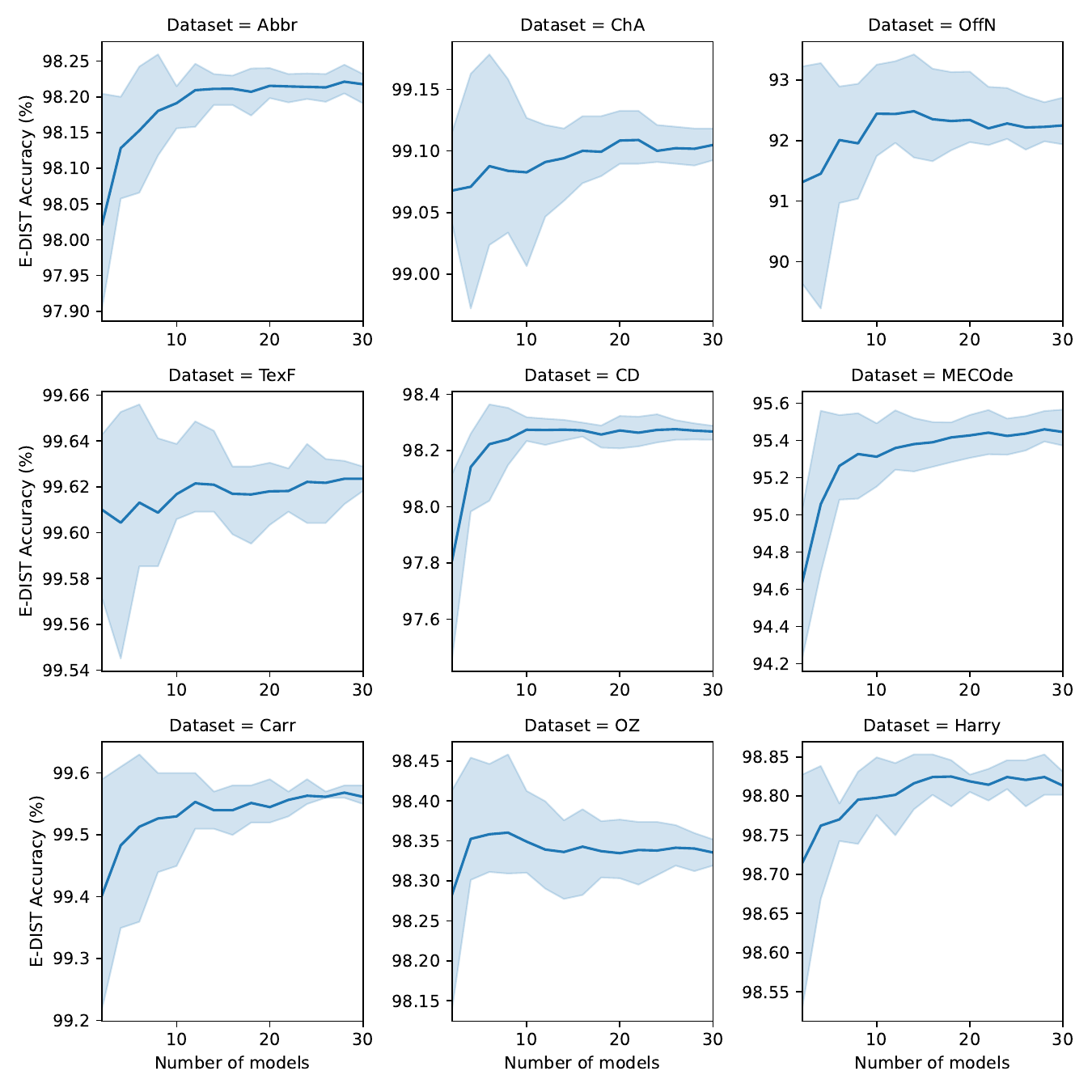}	
	\caption{Effect of number of model instances in E-DIST for each dataset. Note the number of models refers to the total number of model instances in E-DIST, meaning half of that number for each normalization scheme. Hence, the plot starting at two. To take into account the differences in achieved accuracy caused by which model instances are added to the ensemble, the data shown is the result of running the experiment 15 times for each dataset with the order in which the model instances are added being randomized each time. The blue line indicates the average and the shaded blue area shows the range of values. To show the trends for each dataset, only the scale of the \textit{x}-axis is kept constant.}
	\label{fig:ensnummodels}
\end{figure}

In \figurename~\ref{fig:ensnummodels} we show how the absolute accuracy achieved by E-DIST develops depending on how many instances of the DIST model are used. Due to the DIST model's performance being strongly dependent on the data normalization and it being desirable for the end user to not have to choose the most appropriate normalization for their data the E-DIST uses fixation data normalized in two different ways. Since the achieved accuracy depends on which model instances are added to E-DIST, the data shown in \figurename~\ref{fig:ensnummodels} is the result of running the experiment 15 times with the order in which model instances are added being randomized each time.

As can be seen in \figurename~\ref{fig:ensnummodels}, the accuracy for all datasets tends to increase when a higher number of model instances is used in the ensemble. \hl{While the overall effect size is small for all datasets, it is nevertheless interesting that this} trend is particularly pronounced for the two MECOde and OffN datasets which show the biggest difference in performance for the two normalization schemes. Since adding additional model instances to the ensemble has diminishing returns and comes at the cost of being more computationally demanding the E-DIST accuracy reported in Table~\ref{tbl-classicCompare} is based on using three model instances for each normalization scheme, so six instances in total.
\clearpage

\section{Enlarged Figures}
\label{s-sec-larger-figs}

\begin{figure}[ht!]
	\centering
		\centering 
		\includegraphics[width=\textwidth,height=0.8\textheight,keepaspectratio]{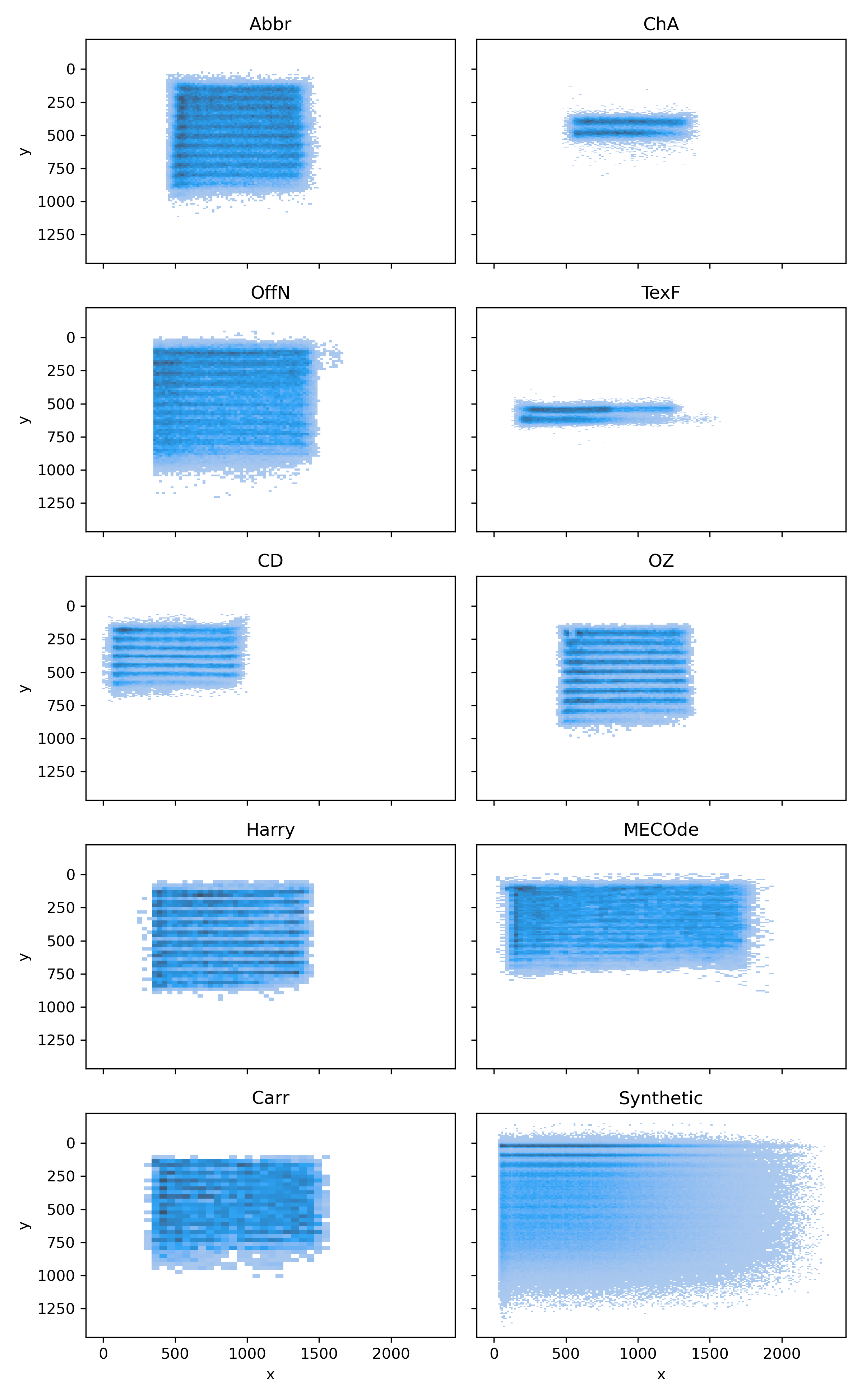}	
		\caption{No normalization with large spread of position and extend.}
		\label{fig:raw_dsets_suppl}
\end{figure}

\begin{figure}[ht!]
		\centering 
		\includegraphics[width=\textwidth,height=0.8\textheight,keepaspectratio]{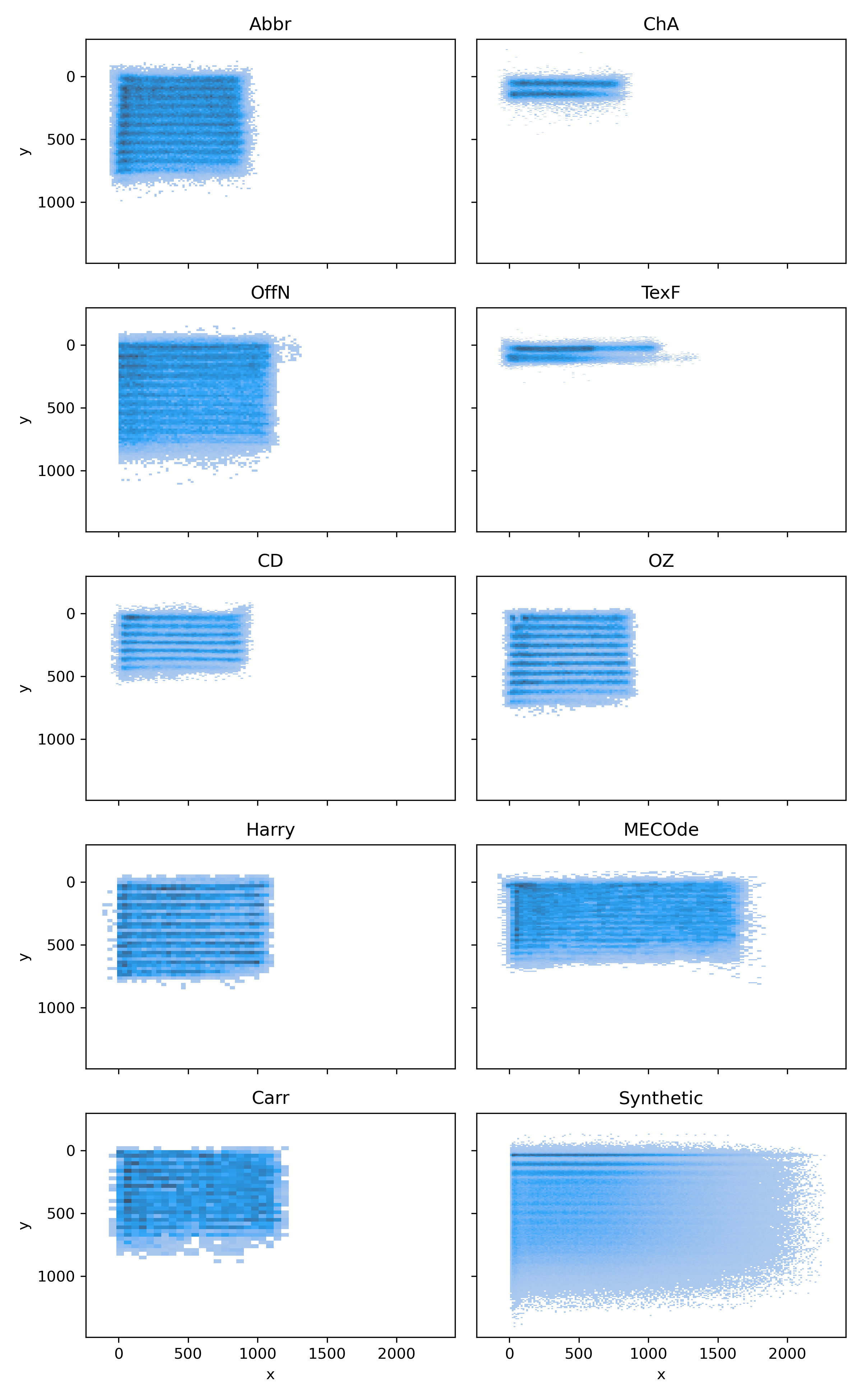}	
		\caption{After only applying xy-norm.}
		\label{fig:xyminsubstracted_dsets_suppl}
\end{figure}

\begin{figure}[ht!]
		\centering 
		\includegraphics[width=\textwidth,height=0.8\textheight,keepaspectratio]{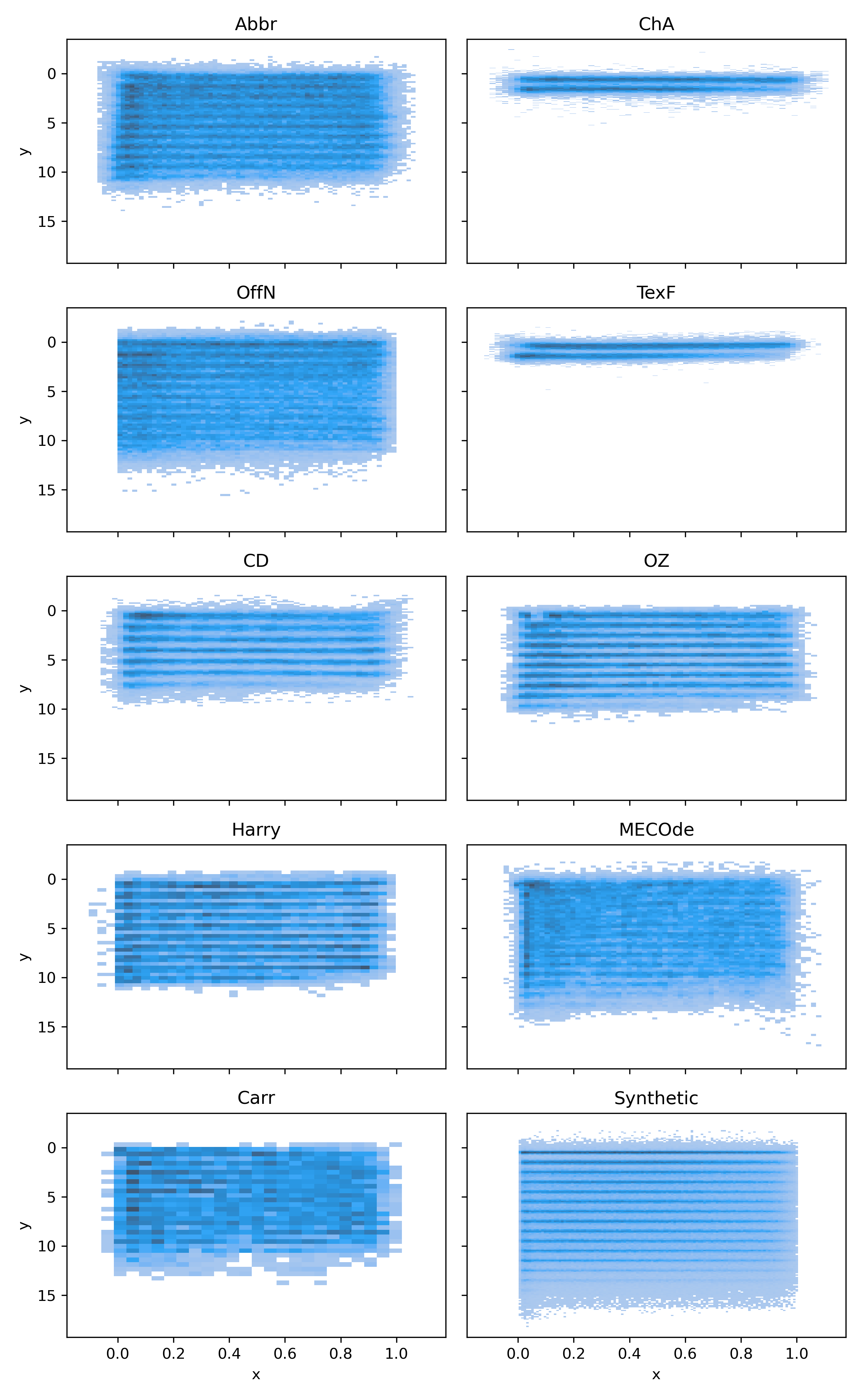}	
		\caption{After applying both xy-norm and lw-norm.}
		\label{fig:xyminsubstractedANDheightwidth_dsets_suppl}
\end{figure}

\clearpage
\bibliographystyle{IEEEtran} 
\bibliography{Eye-Tracking.bib}